\documentclass[final]{cvpr}

\usepackage[square,numbers]{natbib}
\usepackage{algorithm}
\usepackage{algorithmicx}
\usepackage{algpseudocode}

\usepackage{amsmath}
\DeclareMathOperator*{\argmax}{argmax} 
\DeclareMathOperator*{\argmin}{argmin}
\usepackage{times}
\usepackage{epsfig}
\usepackage{graphicx}
\usepackage{amsmath}
\usepackage{amssymb}

\usepackage{amsthm}

\newtheorem{theorem}{Theorem}
\usepackage[english]{babel}
\usepackage{blindtext}
\usepackage{lipsum}  

\usepackage{array}
\usepackage{booktabs}

\usepackage{multirow}

\newcommand{\mulint}{\int_{-\pi}^{\pi}\cdots \int_{-\pi}^{\pi}}

\newcommand{\ejw}{e^{j\omega}}

\newcommand{\w}{\omega}

\newcommand{\Z}{\mathbb{Z}}

\newcommand{\G}{\mathcal{G}}

\newcommand{\oRe}{\operatorname{Re}}

\newcommand{\norm}{\vert\vert}

\newcommand{\yaps}{y_{\text{APS}}}

\newcommand\blfootnote[1]{%
	\begingroup
	\renewcommand\thefootnote{}\footnote{#1}%
	\addtocounter{footnote}{-1}%
	\endgroup
}

\usepackage[pagebackref=true,breaklinks=true,colorlinks,bookmarks=false]{hyperref}

\def\cvprPaperID{10159} 
\def\confYear{CVPR 2021}

\begin{document}

\title{Truly shift-invariant convolutional neural networks}

\author{Anadi Chaman\\
University of Illinois at Urbana-Champaign\\
{\tt\small achaman2@illinois.edu}
\and
Ivan Dokmanić\\
University of Basel\\
{\tt\small ivan.dokmanic@unibas.ch}
}

\maketitle

\begin{abstract}

Thanks to the use of convolution and pooling layers, convolutional neural networks were for a long time thought to be shift-invariant. However, recent works have shown that the output of a CNN can change significantly with small shifts in input---a problem caused by the presence of downsampling (stride) layers. The existing solutions rely either on data augmentation or on anti-aliasing, both of which have limitations and neither of which enables perfect shift invariance. Additionally, the gains obtained from these methods do not extend to image patterns not seen during training. To address these challenges, we propose adaptive polyphase sampling (APS), a simple sub-sampling scheme that allows convolutional neural networks to achieve 100\% consistency in classification performance under shifts, without any loss in accuracy. With APS, the networks exhibit perfect consistency to shifts even before training, making it the first approach that makes convolutional neural networks truly shift-invariant. \blfootnote{Code available at \url{https://github.com/achaman2/truly_shift_invariant_cnns}.}



\end{abstract}

\section{Introduction}



The output of an image classifier should be invariant to small shifts in the image. For a long time, convolutional neural networks (CNNs) were simply assumed to exhibit this desirable property \cite{lecun2015deep_nature,doi:10.1162/neco.1989.1.4.541,lecun_1990,lecun_gradient_based_learning}. This was thanks to the use of convolutional layers which are shift equivariant, and non-linearities and pooling layers which progressively build stability to deformations \cite{Bietti_mairal_group_invariance,mallat_group_invariant_scattering}. However, recent works have shown that CNNs are in fact not shift-invariant \cite{Azulay_Weiss, Zhang19, landscape_of_spatial_robustness,cnn_absolute_position, Amirul_jia_position_info}. Azulay and Weiss \cite{Azulay_Weiss} show that the output of a CNN trained for classification can change with a probability of $30\%$ with merely a one-pixel shift in input images. Related works \cite{Amirul_jia_position_info, cnn_absolute_position} have also revealed that CNNs can encode absolute spatial location in images: a consequence of a lack of shift invariance.\\

\begin{figure}[t]
	\begin{center}
		\includegraphics[width=1\linewidth]{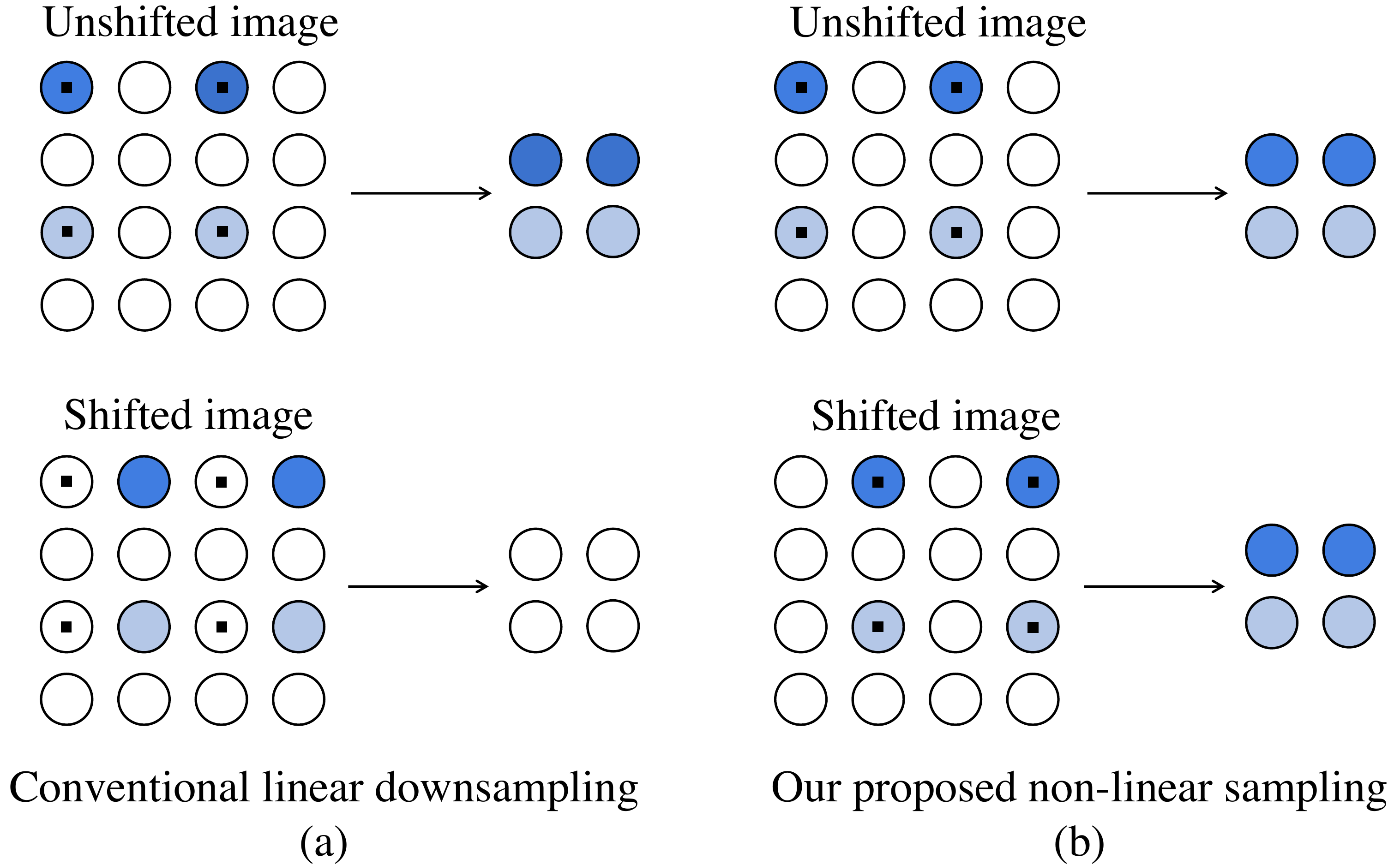}
	\end{center}



\caption{(a) Conventional downsampling is not robust to shifts. It samples image pixels at fixed locations on the grid (shown with small squares). Shifting the image changes pixel intensities located on the fixed grid, resulting in a different subsampled output. (b) By choosing the sampling grid that supports pixels with highest energy, our approach results in shift invariance.}


	\label{fig:downsampling_invariance_culprit}
	
\end{figure}

\begin{figure}[t]
	\begin{center}
		\includegraphics[width=1\linewidth]{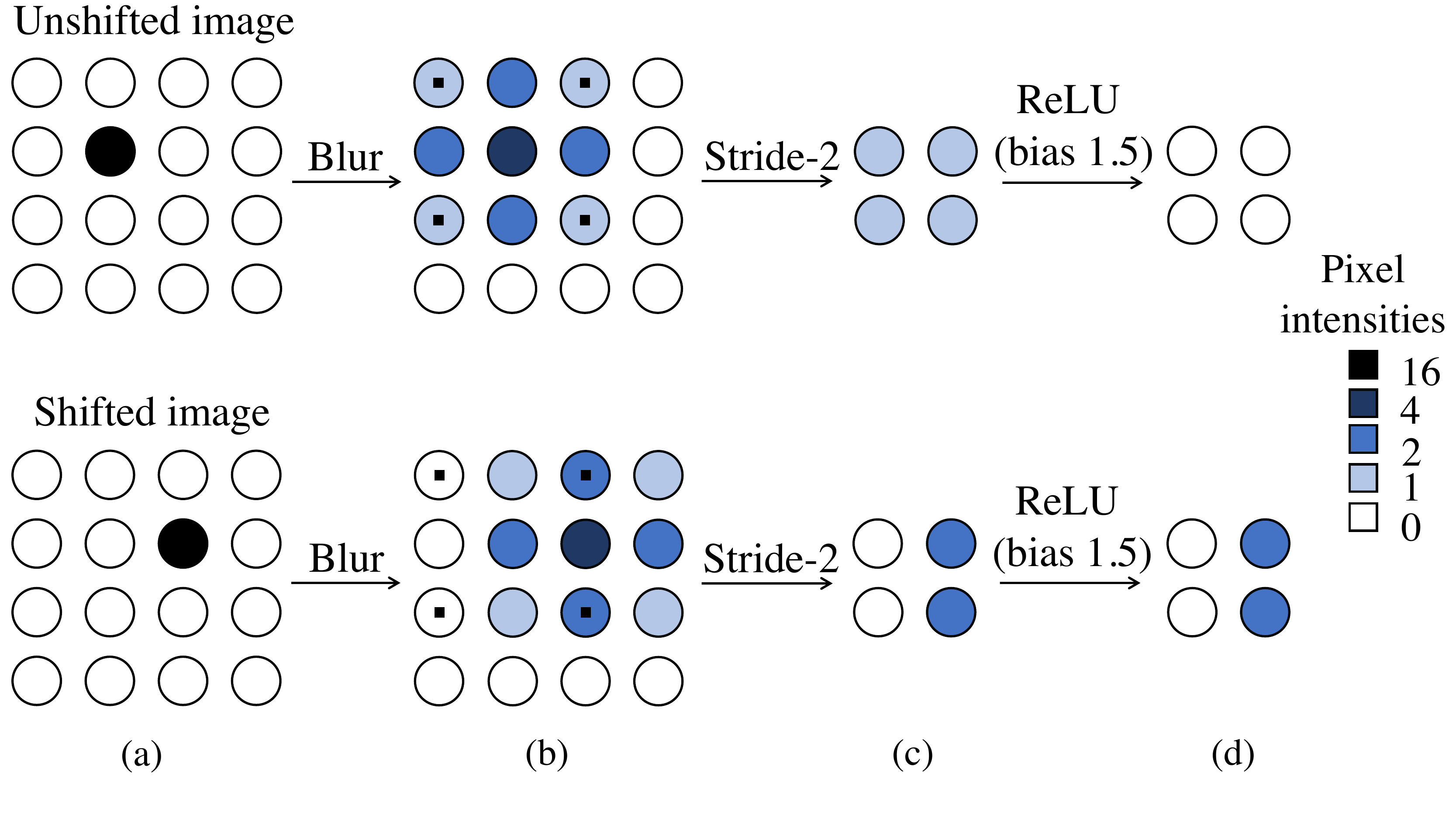}
	\end{center}
	\caption{Anti-aliasing based solutions. (a) An image and its shift. (b)-(c) Blurring spreads sharp features across their neighbouring pixels. For the 2 images, the resulting pixel intensities over the sampling grid become more similar improving shift invariance. (d) Due to minor differences in the subsampled outputs in (c), they are thresholded differently by ReLU and shift invariance is lost. }
	\label{fig:blurpool_and_relu}
	
\end{figure}

One of the key reasons why CNNs are not shift-invariant is downsampling\footnote{Layers like strided max-pooling in CNNs can be regarded as a combination of a dense max-pooling operation followed by downsampling. } \cite{Azulay_Weiss, Zhang19}, or stride, which is a linear operation that samples evenly spaced image pixels located at fixed positions on the grid and discards the rest. As shown in Fig. \ref{fig:downsampling_invariance_culprit}(a), the results of downsampling an image and its shifted version can be significantly different. This is because shifting an image can change the pixel intensities located over the sampling grid. Various measures have been proposed in literature to counter this problem. With data augmentation \cite{simoyan_vgg_2015}, the output of a CNN can be made more robust to shifts by training it on randomly shifted versions of input images \cite{ Azulay_Weiss,Zhang19}. This, however, improves the network's invariance only for image patterns seen during training \cite{Azulay_Weiss}. Anti-aliasing or blurring spreads sharp image features across their neighbouring pixels which improves structural similarity between subsampled outputs of an image and its shifted version (Fig. \ref{fig:blurpool_and_relu}(a)-(c)). One instance of this technique are strided average pooling layers \cite{lecun_1990}. Azulay and Weiss \cite{Azulay_Weiss} showed that anti-aliasing a linear convolutional network with global pooling in the end completely restores shift invariance. Zhang \cite{Zhang19} combined a dense max pooling layer and strided blurred pooling to boost shift invariance and accuracy of classification. While blurring-based methods do improve the network's robustness to shifts, they only achieve partial shift invariance. Their performance is limited by the action of non-linear activation functions such as ReLU \cite{Azulay_Weiss}; we illustrate it in Fig. \ref{fig:blurpool_and_relu}(d). Additionally, anti-aliasing beyond a certain point can result in over-blurring and loss of accuracy \cite{Zhang19}.\\


Taking inspiration from the recent works of Azulay and Weiss \cite{Azulay_Weiss} and Zhang \cite{Zhang19}, we propose adaptive polyphase sampling (APS)---a simple non-linear downsampling scheme that provides substantial improvement in classification consistency over blurring-based methods. In fact, with only mild changes in the padding used by convolutional layers we show that APS enables 100\% consistency in classification performance under shifts without any loss in accuracy, making it the first approach that makes CNNs \textit{truly shift-invariant}. \\

APS achieves the aforementioned performance by addressing the root cause of the problem---the use of a fixed grid by current architectures to subsample an image, when there actually exist multiple sampling grids to choose from (Fig. \ref{fig:downsampling_invariance_culprit}(b)). The key idea used by APS is that shifting an image simply translates its pixels from one grid to the other. Therefore, instead of always choosing a fixed sampling grid, one can select the grid adaptively—for example by choosing the one which supports pixels with the highest energy. The resulting subsampled outputs are then identical for any shift in input. \\

With APS, the achieved boost in shift invariance is completely unaffected by the non-linear ReLU activations (or any other point-wise activations) in the network, and the improved robustness to shifts is consistent regardless of whether the network is tested on in- or out-of-distribution samples. Furthermore, CNNs using APS exhibit complete shift invariance even before training, implying that the invariance prior is truly embedded in the architecture. This prior also improves generalization, thus improving classification accuracy on CIFAR-10 \cite{krizhevsky2009learning} and ImageNet \cite{deng2009imagenet} datasets. APS does not require any additional learnable parameters, can be easily integrated into existing architectures, and also eliminates the need for overblurring the network's feature maps to improve classification consistency with respect to shifts.








\section{Related work}
\label{sec:related_work}

Embedding invariance to shifts in the architecture of neural networks has been studied already in the 1980s \cite{BARNARD1990403,lecun_1990,FUKUSHIMA2003161,lecun_gradient_based_learning}. Robustness of neural networks to spatial and geometric transformations has been quantitatively assessed  \cite{Kanbak_2018_CVPR, goodfellow_measuring_invariances,landscape_of_spatial_robustness,fawzi_are_classifiers_invariant,Azulay_Weiss,ruderman2018pooling, manfredi2020shift} and specialized neural network architectures have been proposed to improve invariance to transformations such as rotation, scaling and arbitrary deformations \cite{Cheng_rotation_invariant_cnn,cyclic_symmetry_rotation_cnn,scale_invariant_cnn,deformable_cnns,Worrall_2017_CVPR,group_equivariant_convolution_networks,steerable_rotation_equivariant_cnn,warped_convolutions,VANNOORD2017583,DBLP:journals/corr/KanazawaSJ14}. Theoretical works based on wavelet filter banks \cite{Bruna_mallat_invariant_scattering,Sifre_Mallat_Rotation_2013_CVPR,mallat_group_invariant_scattering} and multi-layer kernels \cite{Bietti_mairal_group_invariance,bietti_mairal_invariance_stability,conv_kernel_networks} have addressed stability of neural networks to deformations. A body of work has also studied the robustness of CNNs to image corruptions \cite{Hosseini_google_cloud, vasiljevic2016blur_cnn,8038465,geirhos2017comparing,geirhos_generalization_in_humans}. Our focus is rather on restoring shift invariance lost due to subsampling.\\

A parallel line of work is adverserial training, whereby specifically designed perturbations with small norms are added to input images to yield large changes in the network's output \cite{madry2018towards,szegedy2013intriguing,ilyas2019adversarial,explaining_harnessing_adverserial}. We focus on the robustness of networks to shifts which are examples of more naturally occurring transformations. Note that small shifts, despite being imperceptible, may yield images comparably far from the unshifted ones in $l_p$ norms.\\

Data augmentation has been shown to improve robustness and generalization \cite{lecun_gradient_based_learning,pmlr-v15-bengio11b,he_2016_resnet,wu2015deep_image,cutout_regularization,augmix,cubuk_auto_augment}. However, the robustness gains do not carry over to previously unseen transformations and out-of-dataset image distributions \cite{geirhos_generalization_in_humans, Azulay_Weiss}. Engstrom\textit{ et al}. \cite{landscape_of_spatial_robustness} showed that while data augmentation improves robustness to shifts and rotations on average, there still exist transformations that can adversely impact the performance of the trained models.\\

Lack of shift invariance due to downsampling was noted early on by Simoncelli \textit{et al.} \cite{simoncelli_shiftable_multiscale}. They showed that shift invariance in multi-scale convolutional transforms is not possible, and instead proposed the notion of \textit{shiftability}, a weaker form of invariance associated with anti-aliasing. Zhang \cite{Zhang19} showed that both consistency to shifts and accuracy of classification can be improved by combining a dense max pooling layer with anti-aliasing before sampling. Zou \textit{et al.} \cite{zou2020delving} used content-aware anti-aliasing to reduce risk of signal loss from over-blurring. Instead of explicitly anti-aliasing the feature maps, Sundaramoorthi and Wang \cite{translation_insensitive_cnns} showed that by parameterizing convolutional filters with smooth Gauss--Hermite basis functions, CNN classifiers can attain translation insensitivity, a weak form of shift invariance. Azulay and Weiss \cite{Azulay_Weiss}, on the other hand, showed that while anti-aliasing can improve shift invariance, it offers only a partial solution. This is because the improved robustness to shifts is limited by the action of non-linear activations like ReLU,  and  does  not  generalize well to image patterns not seen during training.\\

Removing stride layers from CNNs can restore shift invariance \cite{Azulay_Weiss}. This is indeed the case with networks based on \emph{\`a trous}  convolutions \cite{Yu_2017_CVPR,chen2014semantic,7913730}. Alas, this leads to a dramatic increase in memory and computation requirements, rendering it an impractical strategy for large networks. Shift invariance in convolutional neural networks can also be lost because of boundary and padding effects that arise due to the finite support of input images \cite{cnn_absolute_position,Amirul_jia_position_info,mind_the_pad}. This allows CNNs to encode absolute spatial locations in images.

%
%



\section{Our proposed approach}
\label{sec:our_proposed_approach}
\subsection{Preliminaries}
\label{sec:preliminaries}

\textbf{Shift invariance.} An operation $\G$ is said to be shift invariant if for a signal $x$ and its shifted version $x_s$, $\G(x) = \G(x_s)$. Similarly, it is shift equivariant if $\G(x_s) = (\G(x))_s$. Convolution is an example of a shift equivariant operator. We define $\G$ as sum-shift-invariant if $\sum \G(x_s) = \sum \G(x)$, where the summation is over the pixels of $\G(x)$ and $\G(x_s)$ respectively.\\

Convolutional neural networks for classification contain fully connected layers at the end which are not shift invariant. As a result, any shifts in convolutional feature maps of the final layer can impact the classifier's final output. Global average pooling, popularly used in CNN architectures like ResNet \cite{he_2016_resnet} and MobileNet \cite{howard2017mobilenets}  can solve this problem. These layers reduce the feature map of each channel in the final convolutional layer to a scalar by averaging. Thus, if the convolutional part of the network is sum-shift-invariant, the overall classifier architecture can be made shift invariant. Our analysis in subsequent sections assumes the use of global average pooling layers. \\

\textbf{Polyphase components.} For simplicity, we will consider downsampling of 1-D signals with stride 2. The analysis easily generalizes to images and volumes. Consider a 1-D signal $x_0(n)$ with discrete-time Fourier transform (DTFT) $X_0(\ejw)$, which we will denote by $X_0(\w)$ from hereon. The DTFT of a one-pixel-shifted version $x_1(n) = x_0(n-1)$ is given by $X_0(\w)e^{-j\w}$. Given $x_0(n)$, there are two ways to uniformly sample it with stride 2: we can choose to retain samples at either even or odd locations on the grid. These two possible downsampled outputs denoted by $y_0$ and $y_1$ are called the even and odd polyphase components of $x_0$, and can be expressed as $y_0(n) = x_0(2n)$ and $y_1(n) = x_0(2n-1)$. Notice that the even polyphase component of $x_1$ is the same as the odd counterpart of $x_0$ and vice versa. The downsampled outputs $y_0(n)$ and $y_1(n)$ have DTFTs given by 
\begin{align}
\label{eq:polyphase_0_dtft}
Y_0(\w) =& \text{ }\frac{X_0(\w/2) + X_0(\w/2+\pi)}{2},\\
\label{eq:polyphase_1_dtft}
Y_1(\w) = &\text{ }\frac{(X_0(\w/2) - X_0(\w/2+\pi))e^{-j\w/2}}{2}.
\end{align}
The terms in \eqref{eq:polyphase_0_dtft} and \eqref{eq:polyphase_1_dtft} corresponding to $(\w/2+\pi)$ are called aliased components. They arise when $x_0$ contains high frequencies, and can cause significant degradation of the subsampled outputs. This is traditionally countered by anti-aliasing \cite{oppenheim2001discrete}, a signal processing technique which removes high frequencies in $x_0$ by blurring before sampling.\\

Global average pooling operation on a signal $x_0(n)$ results in its mean and, ignoring a normalizing constant, is equal to $X_0(\w = 0)$.\\

\vspace{-2mm}



\subsection{Key problem with downsampling}
\label{sec:aps_motivation}
Downsampling is used in CNNs to increase the receptive field of convolutions, and to reduce the amount of memory and computation needed for training. With these goals, either of the two polyphase components of a 1-D signal is a `valid' result of downsampling. However, when using conventional linear sampling, current neural network architectures always select the even component, rejecting the odd one. As a result, downsampling $x_0$ and its shifted version $x_1$ always results in different signals $y_0$ and $y_1$ which are highly unlikely to be equal \cite{simoncelli_shiftable_multiscale} or sum-shift-invariant. Indeed, we can see from \eqref{eq:polyphase_0_dtft} and \eqref{eq:polyphase_1_dtft} that $Y_0(0)\neq Y_1(0)$. Anti-aliasing based methods \cite{Zhang19} attempt to improve invariance by promoting similarity between $y_0$ and $y_1$. In particular, they restore sum-shift-invariance by blurring before downsampling, resulting in outputs $y_0^a$ and $y_1^a$ with DTFTs,
\begin{align}
Y_0^{a}(\w) = \frac{X_0(\w/2)}{2},\text{ }
Y_1^{a}(\w) = \frac{X_0(\w/2)e^{-j\w/2}}{2}.
\end{align}
\noindent
The resulting $y_0^a$ and $y_1^a$ do satisfy $Y^a_0(0)=Y^a_1(0)$, i.e., $\sum y_0^a = \sum y_1^a$. This desirable equality, however, is spoiled by the action of ReLU in subsequent layers. As Fig. \ref{fig:blurpool_and_relu}(c) suggests, while $y_0^a$ and $y_1^a$ are similar, they are not identical. Minor differences between the signals become more prominent when they are thresholded by the ReLU non-linearity, resulting in $\sum \mathrm{relu}(y_0^a) \neq \sum \mathrm{relu}(y_1^a)$ (see \ref{sec:relu_explanation} in {\textit{ supplementary material}} for a more formal discussion). One could ask---can increasing the amount of blurring alleviate this problem caused by ReLUs? The answer is \textit{no}. In fact, even ideal low-pass filtering does not help. This is because irrespective of the type of anti-aliasing used, $y_0^a$ and $y_1^a$ always have some differences and, therefore, are thresholded differently by ReLU. \\

While various non-linearities can spoil sum-shift-invariance similar to ReLU, exceptions like polynomial activations do not cause this problem. We state this formally in Theorem \ref{thm:polynomial} (proof in \ref{sec:polynomial_non_lin} in \textit{supplementary material}).



%
%
%
%



\begin{theorem}
	\label{thm:polynomial}
	Given non-linear activation function $g(y) = y^m$ with integer $m>1$, and anti-aliased outputs of sampling $y_0^a$ and $y_1^a$ as defined above, we have
	\begin{equation}
		\sum_{n\in \Z}g(y_0^a)(n) = \sum_{n\in \Z}g(y_1^a)(n).\\
	\end{equation}
	
\end{theorem}

Note that the above discussion and conclusions directly apply to 2-D images, with the difference that instead of 2, there exist 4 polyphase components to choose from.

\begin{figure}[t]
	\begin{center}
		\includegraphics[width=1\linewidth]{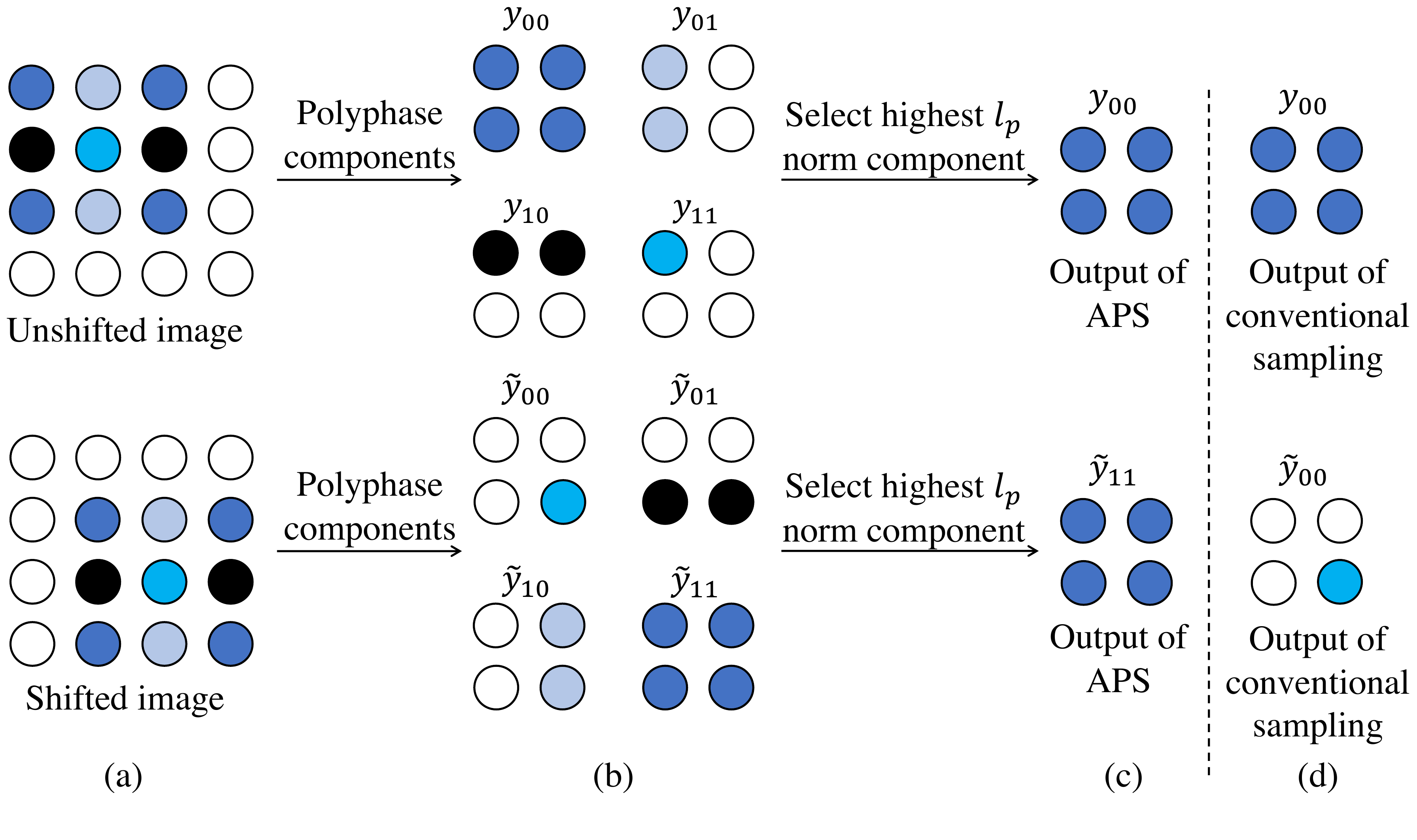}
	\end{center}

\caption{APS on single channel input. (a) Image and its shift. (b) The two images share the same set of polyphase components (with a potential shift between them). (c) By choosing the component with the highest $l_p$ norm, APS returns the same output for both the images. (d) Output of conventional sampling in contrast.}
	\label{fig:aps_illustration}
	
\end{figure}

\subsection{Adaptive polyphase sampling}
\label{sec:aps}
Consider stride-2 subsampling of a single channel image $x$. As shown in Fig. \ref{fig:aps_illustration}(a)-(b), the image can be downsampled along 4 possible grids, resulting in the set of 4 potential candidates for subsampling. We refer to these candidate results of sampling as polyphase components and denote them by $\{y_{ij}\}_{i,j = 0}^1$. Similarly, the polyphase components of a 1-pixel shifted version of $x$, namely $\tilde{x} = x(m-1,n-1)$, are denoted by $\{\tilde{y}_{ij}\}_{i,j = 0}^1$. Notice from Fig. \ref{fig:aps_illustration}(b) that $\{\tilde{y}_{ij}\}$ is just a re-ordered and potentially shifted version of the set $\{y_{ij}\}$. More formally,
\begin{alignat}{2}
\label{eq:polphase_comp_shift_0}
\tilde{y}_{00} =&\text{ }y_{11}(n_1-1, n_2-1), \quad &&\tilde{y}_{10} = y_{01}(n_1, n_2-1),\\
\tilde{y}_{01} =&\text{ } y_{10}(n_1-1, n_2), \quad &&\tilde{y}_{11} = y_{00}(n_1,  n_2).\notag
\end{alignat}


As we saw in Section \ref{sec:aps_motivation}, the key reason why conventional sampling is not shift invariant is that it always returns the first polyphase component of an image as output. This results in $y_{00}$ and $\tilde{y}_{00}$ as subsampled outputs, which from \eqref{eq:polphase_comp_shift_0} are not equal. To address this challenge, we propose adaptive polyphase sampling (APS). The key idea that APS exploits is that $\{y_{ij}\}$ and $\{\tilde{y}_{ij}\}$ are sets of identical\footnote{The images in the two sets could have some shifts between them as well. However, this does not impact shift invariance for networks ending with global average pooling.} but re-ordered images. Therefore, the same subsampled output for $x$ and $\tilde{x}$ can be obtained by selecting a polyphase component from $\{y_{ij}\}$ and $\{\tilde{y}_{ij}\}$ in a permutation invariant manner—for example choosing the one with the highest $l_p$ norm. This is illustrated in Fig. \ref{fig:aps_illustration}(c). APS obtains its output $\yaps$ by using the following criterion with $p=2$.
\begin{gather}
\label{eq:aps_criterion}
\yaps = y_{i_1j_1},\\
\text{where } i_1, j_1 = \argmax_{i,j} \{ \norm y_{ij} \norm_p  \}_{i,j=0}^1. \notag
\end{gather}

\noindent
For reference, conventional sampling returns $y_c = y_{00}$ as the output for $x$. It can be observed that for a shift $n_0>1$ between $x$ and $\tilde{x}$, the resulting subsampled outputs $\yaps$ and $\tilde{y}_{\text{APS}}$ are identical upto a shift $\sim \lceil\frac{n_0}{2}\rceil$, making the operation sum-shift-invariant. Additionally, note that since we did not use blurring in \eqref{eq:aps_criterion}, $\yaps$ will contain aliased components if $x$ has high frequencies. This indicates that while anti-aliasing has been shown to improve shift invariance \cite{Zhang19, Azulay_Weiss}, it is not strictly necessary.
\\

\begin{figure}[t]
	\begin{center}
		\includegraphics[width=1\linewidth]{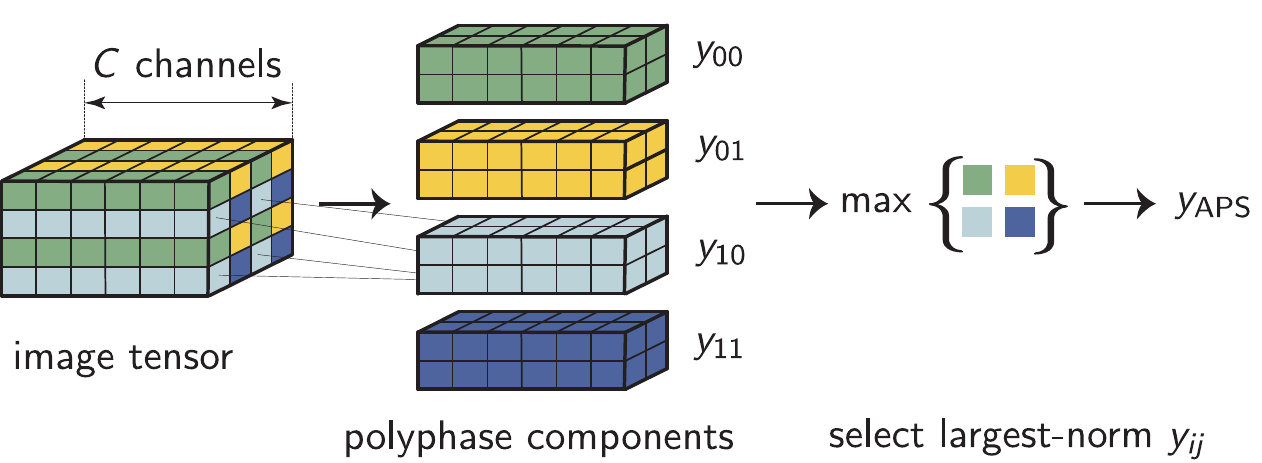}
	\end{center}
	\caption{APS on multi-channel image tensor. The four polyphase components for a multi-channel input are constructed by gathering the components of the individual channels. } 
	\label{fig:multichannel_aps}
	
\end{figure}

When $x$ is an image with $C$ channels given by $x = (x_k)_{k = 1}^C$, we define its polyphase components $\{y_{ij}\}_{i,j=0}^1$, by gathering the respective components for all channels, as shown in Fig. \ref{fig:multichannel_aps}. In particular, if we assume each channel $x_k$, to have components $\{x_{k, ij}\}_{i,j=0}^1$, then for $i,j \in \{0, 1\}$,
\begin{equation}
y_{ij} = (x_{k,ij})_{k=1}^C.
\end{equation}
\noindent
The output of subsampling $x$ using APS, denoted by $\yaps$, can then be obtained similar to \eqref{eq:aps_criterion}. The above method can be extended to a general stride $s$, in a straightforward manner by norm maximization over $s^2$ polyphase components. The overall approach is summarized in Algorithm \ref{alg:aps_algo}.\\

It is possible for APS to downsample inconsistently if two polyphase components in the set $\{y_{ij}\}_{i,j=0}^1$ have equal norms. However, in theory, assuming images to be drawn from a continuous distribution, this is an event with probability zero. It is also very less likely in practice based on our experiments. Note that it is for simplicity that APS chooses polyphase components using norm maximization. If needed, one could choose the components using more sophisticated permutation invariant ways as well.

%
%



\subsubsection{Restoring shift invariance with APS}
\label{sec:restoring_invariance}

Downsampling an image and its shifted version with APS results in outputs that are exactly alike (up to a shift). Therefore, unlike the case with blurring, the action of \textit{any} point-wise non-linearity on these outputs continues to yield identical signals. As a result, for a network containing convolutions, non-linear activations and APS layers, feature maps obtained from an image and its shift are either identical or shifted versions of each other, in all layers. When followed by global average pooling, this results in exactly identical outputs at the end and, therefore, \textit{perfect shift invariance}.

\subsection{Impact of boundary effects on shift invariance}
\label{sec:boundary_effects}

%

While training CNNs to be shift invariant via data augmentation, a standard practice is to show the networks randomly shifted crops of images in the training set. The shifted images obtained this way have minor differences near their boundaries. After each layer these differences are amplified and propagated across the whole image to the point that shift invariance is lost even in the absence of downsampling. One way to resolve this problem is by padding images with enough zeros, though at the expense of additional computational and memory requirements. \\

To separate the two sources of loss in shift invariance---downsampling and boundary effects---while avoiding the extra overhead, we use circular padded convolutions and shifts in our experiments \cite{Zhang19}. We show that with circular padding, CNNs with APS yield 100\% classification consistency to shifts on CIFAR-10 and ImageNet datasets. We then train and evaluate the networks with standard padding and random crop based shifts as well, to still observe superior performance of APS over other approaches. Note that while circular padding helps reduce memory requirements, it can result in additional boundary artifacts when sampling odd sized images. However, even with these artifacts, APS results in better robustness to shifts as compared to prior methods (section \ref{sec:circ_shifted_images_odd} in \textit{supplementary material}).

\subsection{Combining APS with anti-aliasing }
\label{sec:aps_blur_combined}

We saw in Section \ref{sec:aps} that APS can achieve perfect shift invariance without blurring the feature maps. While anti-aliasing is not strictly needed for shift invariance, it is still a useful tool to use before downsampling. This is because, as discussed in Section \ref{sec:preliminaries}, it reduces signal degradation caused by aliased components during sampling. Hence, combining APS with anti-aliasing can help us in reaping the advantage of additional improvements in classification accuracy. This can be done by slightly blurring the feature maps before downsampling them with APS.

\begin{algorithm}[t]
	\begin{algorithmic}[1]
		\State \textbf{Input}: An image $x= \{x_k\}_{k=1}^C$ with $C$ channels.
		\State For $i,j\in\{0, 1, \ldots s-1\} $, polyphase components:

		$\{y_{ij}\}= x(sn_1+i, sn_2+j) = \{x_{k,ij}\}_{k=1}^C.$
		
		\State \textbf{Output}: $\yaps = y_{i_1j_1}$
		
		$\text{where } i_1, j_1 = \argmax_{i,j} \{ \norm y_{ij} \norm_p  \}_{i,j=0}^{s-1}$
		
	\end{algorithmic} 
	\caption{Adaptive Polyphase Sampling with stride s}
	\label{alg:aps_algo}
\end{algorithm}

\begin{table*}[t]
	\centering
	\begin{tabular}{c|*4c| *4c }
		\toprule
		\multicolumn{5}{c|}{Consistency}  & 	\multicolumn{4}{c}{Accuracy (unshifted images)}     \\
		\hline
		Model & ResNet-20   & ResNet-56 & ResNet-18  &ResNet-50 &   ResNet-20  & ResNet-56  & ResNet-18  &ResNet-50     \\
		\hline
		Baseline  &  90.83\% &   91.89\%  &    90.88\%  &   88.96\%    &  89.76\%    &   91.40\%    &    91.96\%    &     90.05\%  \\
		
		APS  &   \textbf{100\%}& \textbf{100\%}    &  \textbf{100\%}    &  \textbf{100\%}     &  90.88\%    &   92.66\%    &   93.97\%     &  94.05\%   \\
		\hline 
		
		LPF-2    &  94.68\% &    94.44\% &   95.06\%   &    92.47\%   &  90.99\%    &  92.07\%     &    93.47\%    &  91.61\%   \\
		
		APS-2   &  \textbf{100\%} &  \textbf{100\%}   &    \textbf{100\%}  &   \textbf{100\%}    &  91.69\%    &   92.28\%    &    94.38\%    &  \textbf{94.27\%}   \\
		\hline
		
		LPF-3   & 95.23\%  &  95.07\%   &   97.19\%   &   95.63\%    &  91.01\%    &  92.24\%     &   94.01\%     &   93.65\%  \\
		
		APS-3  & \textbf{100\%}  & \textbf{100\%}    &   \textbf{100\%}   &   \textbf{100\%}    &  \textbf{91.78\%}    &   92.72\%    &   \textbf{94.53\%}     &  93.80\%   \\ 		
		\hline
		
		LPF-5   &  96.53\% &   96.90\%  &  98.19\%    &   97.38\%    &  91.56\%    &  \textbf{92.98\%  }   &   94.28\%     &   94.12\%   \\
		
		APS-5   &  \textbf{100\%} &  \textbf{100\%}   &  \textbf{100\%}    &  \textbf{100\%}     &   91.75\%   &    92.93\%   &   94.48\%     &   94.07\%  \\
		
		\bottomrule
		
	\end{tabular}
	
	\vspace{2pt}
	\caption{Classification consistency and accuracy (on unshifted images) evaluated on CIFAR-10 test set with ResNet models using blurring (LPF) and APS based downsampling. Circular padding was used in convolutional layers and circular shifts were used for consistency evaluation. The models were not shown any shifted images during training.} 
	\label{tab:circular_shift_cifar}
\end{table*}

\section{Experiments}

We evaluate the performance of APS on CIFAR-10 \cite{krizhevsky2009learning} and ImageNet \cite{deng2009imagenet} datasets. For CIFAR-10, a 0.9/0.1 training/validation fractional split is used over the 50k training set with the final results reported over the test set of size 10k. For models trained on ImageNet, we report the best results evaluated on 50k validation set. APS is compared with 2 types of subsampling methods: (i) BlurPool (anti-aliased sampling) \cite{Zhang19}, and (ii) conventional subsampling, which we regard as baseline. We also evaluate models that combine APS and blurring. Filters of size $2\times 2$, $3\times 3$ and $5\times 5$ have been used for anti-aliasing. The experiments are performed on different variants of the ResNet architecture \cite{he_2016_resnet} whose standard stride layers are replaced by the above subsampling modules. ResNet model embedded with filter size $j \times j$ is denoted by ResNet-LPF$j$. Similarly, ResNet-APS$j$ denotes models containing APS and blur filter of size $j$.  ResNet-18 is used as a running example in our experiments. We compare the models on four metrics.

\begin{itemize}
	
	\item \textbf{Classification consistency on test set.} Likelihood of assigning an image and its shift to the same class.
	
	\item \textbf{Classification accuracy on unshifted test set.} The top-1 accuracy evaluated on unshifted images of the test dataset.

	\item \textbf{Invariance to out-of-distribution image patterns.} We measure classification consistency of the trained networks over image patterns not seen during training.
	 	
	\item \textbf{Stability of convolutional feature maps to small shifts.} We measure the extent to which a 1-pixel shift in input changes the feature maps inside convolutional neural networks. With $y_l$ and $\tilde{y}_l$ as the feature maps for image $x$ and its shift $\tilde{x}$ at depth $l$ in a CNN, we use a shift compensated error $\delta(y_l, \tilde{y}_l)$ as the stability measure. It is defined as	
	\begin{gather}
	\label{eq:stability}
	\delta (y_l, \tilde{y}_l)= | \tilde{y}_l - T_j(y_l) |^2 , \\
	\text{where } j = \argmin_{j_1\in S} \| \tilde{y}_l - T_{j_1}(y_l)  \|_2,\notag
	\end{gather}
	$|\cdot|^2$ represents the squared magnitude function, $T_j$ is an operator that shifts $y_l$ by $j\in S$, and $S$ is a set of all one pixel translations.

\end{itemize}

\noindent
To separate the impact of boundary effects and downsampling on shift invariance, we first implement CNNs with circular padding and evaluate consistency on CIFAR-10 and ImageNet with random circular shifts up to 3 and 32 pixels respectively. All networks are trained with random horizontal flips and without any shifts unless mentioned otherwise. Models trained with random shifts are labelled DA. See \textit{supplementary material} for more details on implementation.

\begin{table*}[t]
	\centering
	\begin{tabular}{c|c |c |c |c |c |c  |c | c   }
		\toprule
		& Baseline & APS & LPF-2&   APS-2 & LPF-3&   APS-3&  LPF-5&   APS-5 \\
		\hline
		Consistency&  80.39\%  &  \textbf{100\%} &  84.35\%   & \textbf{100\%}  & 86.54\%     &   \textbf{99.996\%}   & 87.88\% & \textbf{99.98\%}      \\
		\hline
		Accuracy (unshifted images) &   64.88\%   &   67.05\%    &   67.03\%   &   \textbf{67.60\%} & 66.96\%    &  67.43\%  & 66.85\% & 67.52\%   \\
		\bottomrule

	\end{tabular}
	
	\vspace{2pt}
	\caption{Classification consistency and accuracy (on unshifted images) on ImageNet validation set obtained with ResNet-18 models containing circular padded convolutions and different subsampling modules. Networks were trained without any shifts during training.}
	\label{tab:imagenet_circular_no_aug}
\end{table*}

 


\begin{table*}[t]
	\centering
	\begin{tabular}{c|*4c| *4c }
		\toprule
		\multicolumn{5}{c|}{Consistency}  & 	\multicolumn{4}{c}{Accuracy (unshifted images)}     \\
		\hline
		Model & ResNet-20 & ResNet-56 & ResNet-18& ResNet-50   &   ResNet-20 &ResNet-56 & ResNet-18   & ResNet-50     \\
		\hline
		Baseline   & 88.89\%    &    91.43\%     &    90.81\%     &    90.42\%     &    90.06\%       &     91.14\%     &    91.60\%       &     91.46\%      \\       
		APS       &    92.51\%     &   94.04\%      &    95.12\%     &    95.21\%     &  91.49\%        &    92.89\%       &   93.88\%      &    93.63\%   \\       
		\hline
		LPF-2       &    91.60\%     &   92.30\%      &     94.10\%    &   91.81\%        &   90.67\%       &    91.80\%       &   93.25\%      &   91.93\%  \\       
		APS-2      &  93.62\%       &  94.51\%       &   95.82\%      &     96.08\%      &   91.44\%       &    \textbf{93.04\%} &  93.63\%       &    94.54\%  \\       
		\hline
		LPF-3     &  91.84\%       &   92.63\%      &    94.82\%     &   93.58\%        &    91.73\%      &  92.35\%     &  94.15\%       &  92.66\%  \\       
		APS-3     &   93.45\%      &    94.59\%     &    95.58\%     &    \textbf{96.23\%}       &    91.70\%      &    93.02\%    &  94.31\%       &   \textbf{94.66\%}   \\       
		\hline
		LPF-5       &   93.46\%      &   93.53\%      &   95.13\%      &    94.70\%       &   91.51\%       &     92.55\%      &   94.22\%      &   93.61\%   \\       
		APS-5      &    \textbf{94.08\% }    &    \textbf{94.61\% }    &   \textbf{96.25\%}      &    96.01\%       &    \textbf{92.03\% }     &    92.97\%       &   \textbf{94.81\% }     &    94.37\%   \\

		\bottomrule
		
	\end{tabular}

	\vspace{2pt}
	\caption{Classification consistency and accuracy (on unshifted images) obtained from ResNet models containing APS and blur based subsampling modules. Standard zero padded convolutions were used in the networks and random-crop based shifts were used for consistency evaluation. Networks were trained without any shifts during training.} 
	
	\label{tab:random_crop_cifar_result}
\end{table*}

\begin{figure}[t]
	\begin{center}
		\includegraphics[width=0.92\linewidth]{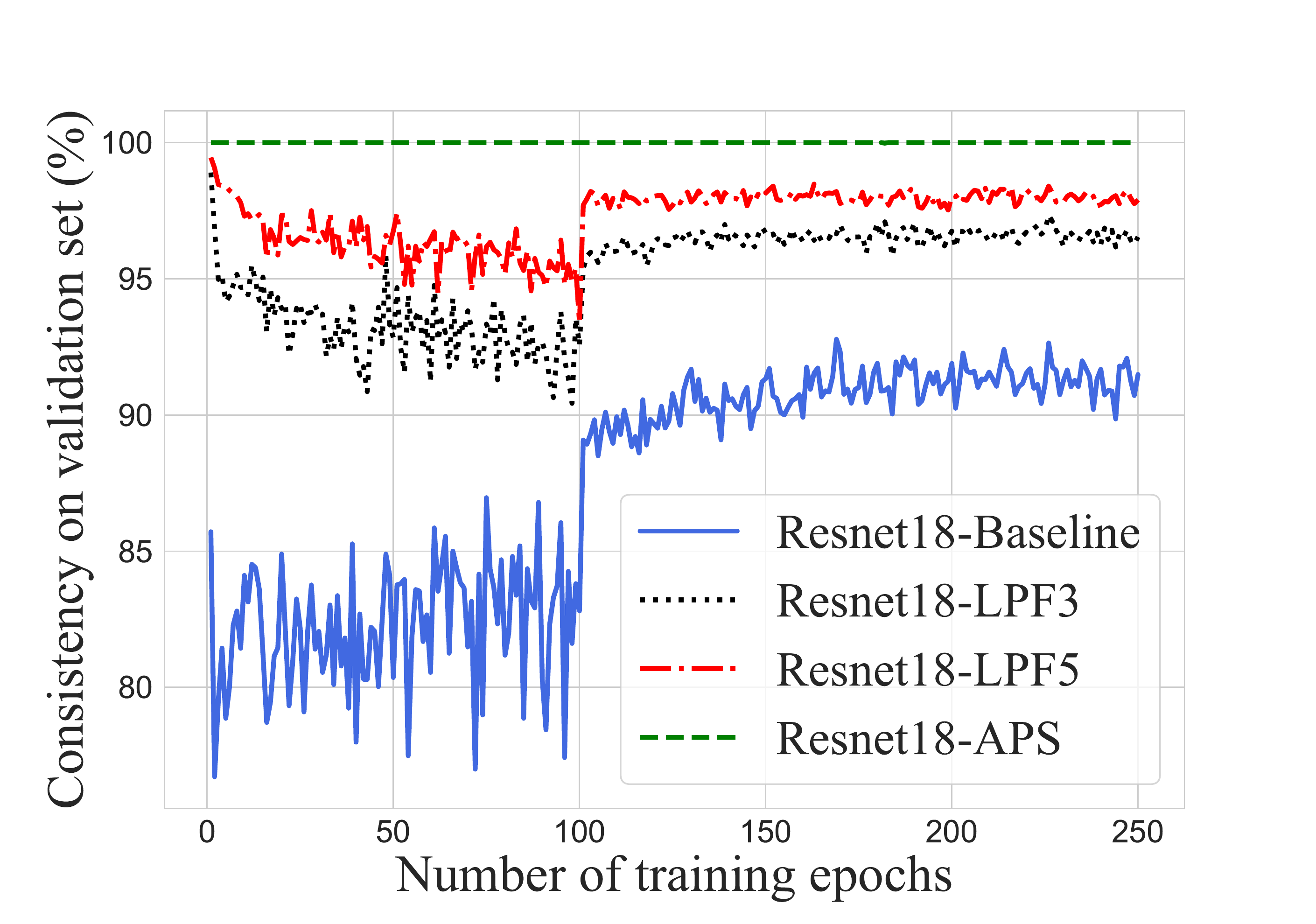}
	\end{center}
	\caption{Classification consistency evaluated after each epoch on the CIFAR-10 validation set for 4 variants of Resnet-18: (i) Baseline, Blurpool with filter size (ii) 3 and (iii) 5, and (iv) APS. } 
	\label{fig:val_consistency_vs_epochs}
	
\end{figure}

\subsection{Classification consistency and accuracy on test test}
\label{sec:classification_consistency_and_acc_test}
We first evaluate 4 ResNet architectures, namely ResNet-20, 56, 18 and 50, with different downsampling modules on CIFAR-10 dataset. Originally used in \cite{he_2016_resnet} for CIFAR-10 classification, ResNet-20 and 56 are small models that downsample twice with stride 2 and contain $\{16, 32, 64\}$ filters in different layers. ResNet-18 and 50 on the other hand, downsample thrice with stride 2 and contain $\{64, 128, 256, 512\}$ filters. Table \ref{tab:circular_shift_cifar} shows consistency and accuracy of the models trained with circular padding. As expected, all networks containing APS modules exhibit perfect robustness to shifts evident from $100\%$ classification consistency. Note that this is despite training the networks without showing any shifted versions of images. In contrast, the baseline ResNet-18 model is consistent $90.88\%$ times, whereas its anti-aliased variants LPF-2, 3 and 5 show consistencies of $95.06\%$, $97.19\%$, $98.19\%$ respectively. Similar to Zhang \cite{Zhang19}, we also observe increase in classification accuracy on unshifted test images with improving shift invariance. For instance, APS increases the accuracy of baseline ResNet-18 from $91.96\%$ to $93.97\%$. We also observe that for a given blur filter size, accuracy obtained by combining APS and anti-aliasing is typically higher than the case with only blurring. As per Section \ref{sec:aps_blur_combined}, we believe this to be a result of combining the perfect shift invariance prior from APS, and anti-aliasing's ability to reduce signal degradation during sampling. \\


To understand the role of learned model weights on robustness to shifts, we compare how classification consistency on CIFAR-10 validation set varies while training ResNet-18 with the different sub-sampling methods. Fig. \ref{fig:val_consistency_vs_epochs} shows that unlike the baseline and anti-aliased models, the validation consistency for APS is $100\%$ throughout training. In fact, we observe perfect consistency in models with APS even before training, implying that APS truly embeds shift invariance into the CNN architecture. \\

We similarly compare the different downsampling methods on ImageNet with ResNet-18 architecture. The results are shown in Table \ref{tab:imagenet_circular_no_aug}. As expected, APS outperforms other methods with higher accuracy and near $100\%$ consistency. Note that the minuscule fall in consistency for APS-3 and 5 occurred due to 2 and 10 images respectively (out of 50k) having polyphase components with the same $l_2$ norm. If needed, this rare occurrence can be avoided by using more robust polyphase component selection methods\footnote{For eg., it is much less likely for two images to have the same $l_p$ and $l_q$ norms for $p\neq q$. One could therefore maximize a sum of two norms to further reduce the likelihood of inconsistent sampling.}.\\



%


\vspace{-3pt}
\textbf{Boundary effects.} As discussed in Section \ref{sec:boundary_effects}, evaluating consistency with random crop based shifts can result in shift invariance loss even in the absence of downsampling. Here, we investigate the impact of these boundary effects on the performance of APS. ResNet models with standard zero padding and different subsampling modules are trained and evaluated on CIFAR-10 dataset. For consistency evaluation, images are padded with zeros of size 3 on all sides, and a crop of size $32\times32$ is randomly chosen. Results in Table \ref{tab:random_crop_cifar_result} reveal that for a given blur filter size, combining APS with anti-aliasing consistently provides better robustness and accuracy compared to blurring alone. In fact, in most cases, the consistency boost provided by APS with no anti-aliasing is still higher than the models that only use blurring.


\subsection{Shift invariance on out-of-distribution images}
\label{sec:out_of_dist_exps}
\begin{figure}[t]
	\begin{center}
		\includegraphics[width=1\linewidth]{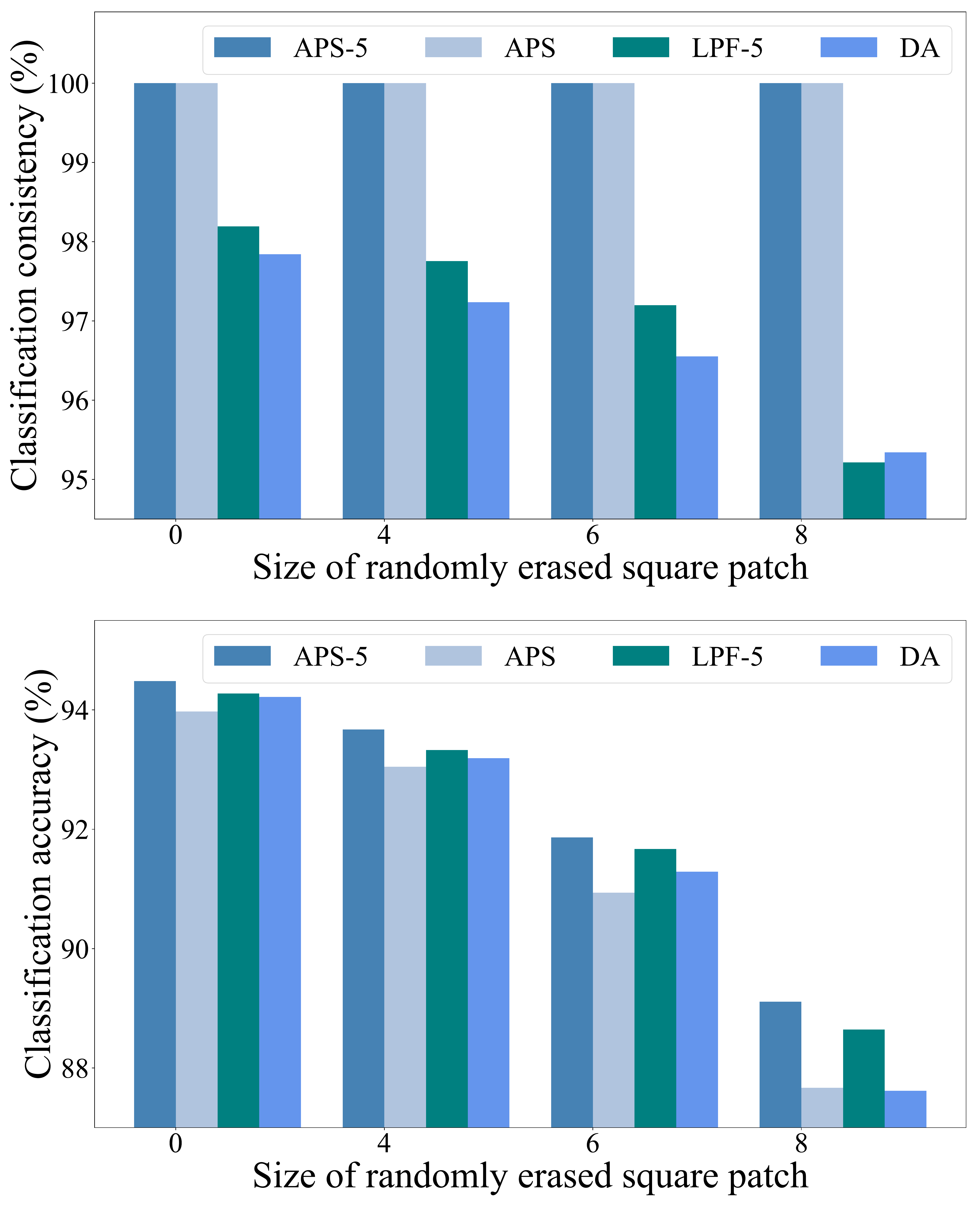}
	\end{center}
	\caption{Classification consistency and accuracy on CIFAR-10 test images with randomly erased square patches. Unlike APS, models with data augmentation and blurring alone lose robustness to shifts as the images move away from training distribution.} 
	\label{fig:random_erasure_consistency_check}
	
\end{figure}

Azulay and Weiss \cite{Azulay_Weiss} showed that robustness to shifts achieved via data augmentation and anti-aliasing gets worse when the trained models are evaluated on images that differ substantially from the training distribution. In our experiments, we make a similar observation. On clean CIFAR-10 images, we train 4 variants of ResNet-18: model with (i) APS, (ii) LPF-5, (iii) APS + blur (APS-5), and (iv) model with vanilla subsampling but trained with random circular shifts of training set (referred to as DA). The trained networks are then evaluated for consistency on test images with small patches of pixels randomly erased from different locations.  Fig. \ref{fig:random_erasure_consistency_check} shows that anti-aliased and data augmentation based models progressively lose robustness to shifts with increasing size of erased patches. On the other hand, models containing APS remain $100\%$ shift invariant. In addition, the network with APS and blurring combined exhibits highest accuracy for all sizes of erasures. \\

We also evaluate these networks on a vertically flipped version of CIFAR-10 test set and observe similar results. Table \ref{tab:flipped_out_of_data} shows that, unlike data augmentation and blurring, models based on APS continue to exhibit 100\% consistency to shifts. Since, vertically flipping the dataset semantically pushes them further away from training set, all the models are expected to show poor accuracy. However, despite that, we observe APS-5 to show higher classification accuracy than other models.




\begin{table}[t]
	\centering
	\begin{tabular}{c|*2c|*2c}
		\toprule
		\multicolumn{3}{c|}{Consistency} & \multicolumn{2}{c}{Accuracy (unshifted)}\\
		\hline
		Models  &  Unflipped & Flipped & Unflipped & Flipped\\
		\hline
		APS-5    &  \textbf{100\% }  &   \textbf{100\%}     &    \textbf{94.48\% }    &    \textbf{47.55\%}     \\
		APS    & \textbf{100\%  } & \textbf{ 100\%  }    &    93.97\%     &    44.79\%     \\
		LPF-5  &  98.19\%  &   89.21\%   &  94.28\%   &   46.21\%    \\
		DA	 &  97.84\%  &  84.94\%  &  94.22\%   &  44.97\%  \\
		\bottomrule

	\end{tabular}
	
	\vspace{2pt}
	\caption{Classification consistency and accuracy evaluated on vertically flipped CIFAR-10 test dataset. Unlike the case with data augmentation and blurring, models with APS continue to remain shift invariant on vertically flipping the dataset. }
	\label{tab:flipped_out_of_data}
\end{table}

%
%
%
%
%

\subsection{Stability of internal convolutional feature maps to small shifts}

\begin{figure}[t]
	\begin{center}
		\includegraphics[width=1.01\linewidth]{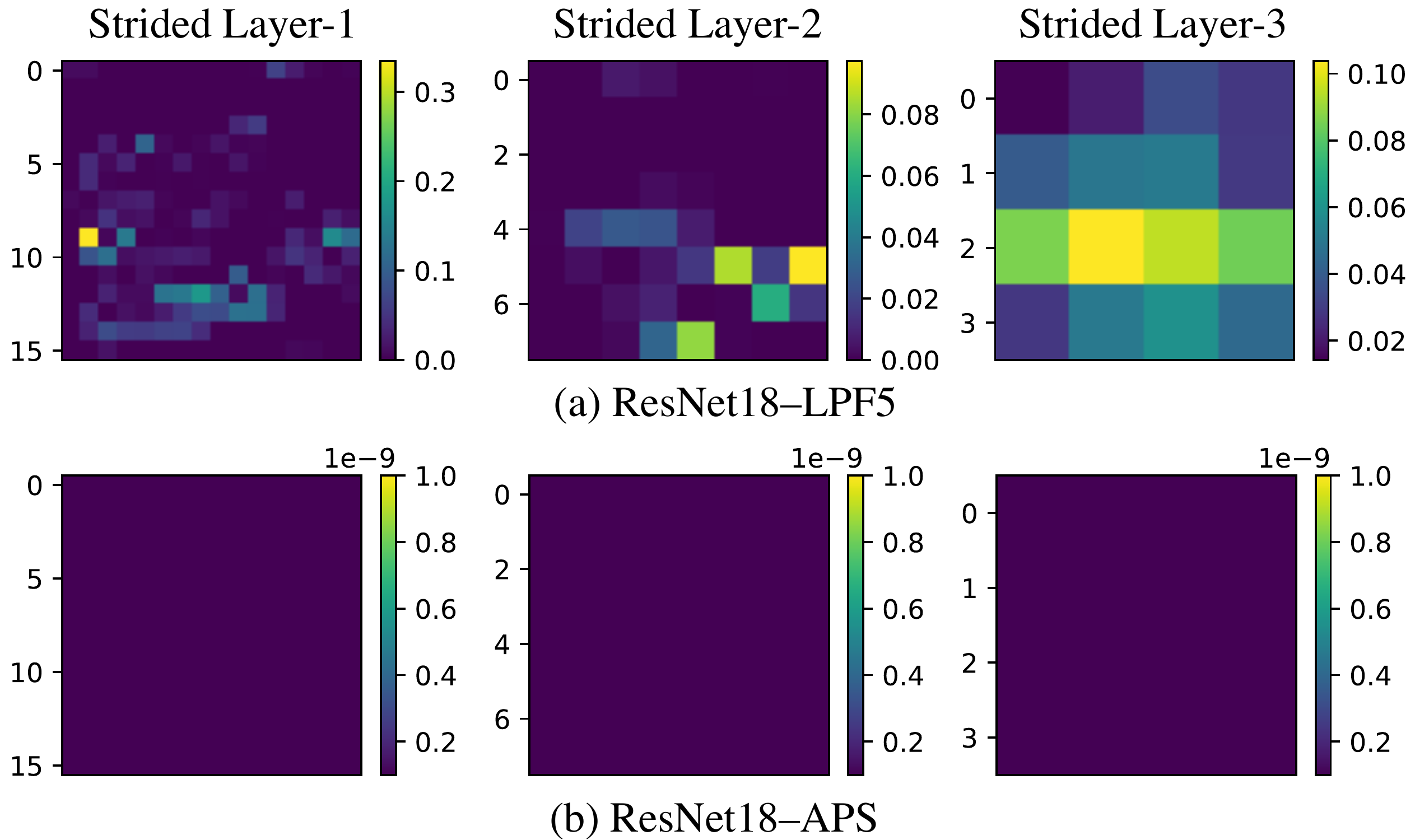}
	\end{center}
	\caption{Stability of internal feature maps. A 1-pixel shift in input results in small differences in the feature maps for ResNet-18 with LPF-5 sampling. However, shift invariant nature of APS results in perfect stability in the feature maps of ResNet18-APS. } 
	\label{fig:conv_stability_lpf_aps}
	
\end{figure}

We compare the impact of diagonally shifting an input image by 1-pixel on the feature maps of ResNet-18 models containing LPF-5 and APS modules. We compute feature maps for a CIFAR-10 test image and its shift, and compare them using shift compensated error $\delta$ from \eqref{eq:stability}. Fig. \ref{fig:conv_stability_lpf_aps} shows the errors for feature maps from the last 3 residual layers of the models (stride-2 sampling used in each layer). For each layer, we plot the errors for channels with the highest energy. The results indicate that while feature maps of LPF-5 model develop minor differences due to shift in input, the output of APS is completely stable.

\section{Conclusion}
\label{sec:conclusion}

Convolutional neural networks lose shift invariance due to subsampling (stride). We address this challenge by replacing the conventional linear sampling layers in CNNs with our proposed adaptive polyphase sampling (APS). A simple non-linear scheme, APS is the first approach that allows CNNs to be truly shift invariant. We show that with APS, the networks exhibit $100\%$ consistency to shifts even \textit{before training}. It also leads to better generalization performance, as evident from improved classification accuracy.

\section*{Acknowledgement}
We thank Konik Kothari for useful discussions on the work presented in this paper. This research was supported by the European Research Council Starting Grant SWING, no. 852821. Numerical experiments were partly performed at sciCORE (\href{http://scicore.unibas.ch/}{http://scicore.unibas.ch/}) scientific computing center at University of Basel. We also utilized resources supported by the National Science Foundation’s Major Research Instrumentation program, grant \#1725729, as well as the University of Illinois at Urbana-Champaign.

{\small
\bibliographystyle{ieee_fullname}

\bibliography{egbib}

\begin{thebibliography}{10}\itemsep=-1pt

\bibitem{mind_the_pad}
Bilal Alsallakh, Narine Kokhlikyan, Vivek Miglani, Jun Yuan, and Orion
  Reblitz-Richardson.
\newblock Mind the pad--cnns can develop blind spots.
\newblock {\em arXiv preprint arXiv:2010.02178}, 2020.

\bibitem{Azulay_Weiss}
Aharon Azulay and Yair Weiss.
\newblock Why do deep convolutional networks generalize so poorly to small
  image transformations?
\newblock {\em Journal of Machine Learning Research}, 20(184):1--25, 2019.

\bibitem{BARNARD1990403}
Etienne Barnard and David Casasent.
\newblock Shift invariance and the neocognitron.
\newblock {\em Neural Networks}, 3(4):403 -- 410, 1990.

\bibitem{pmlr-v15-bengio11b}
Yoshua Bengio, Frédéric Bastien, Arnaud Bergeron, Nicolas
  Boulanger–Lewandowski, Thomas Breuel, Youssouf Chherawala, Moustapha Cisse,
  Myriam Côté, Dumitru Erhan, Jeremy Eustache, Xavier Glorot, Xavier Muller,
  Sylvain~Pannetier Lebeuf, Razvan Pascanu, Salah Rifai, François Savard, and
  Guillaume Sicard.
\newblock Deep learners benefit more from out-of-distribution examples.
\newblock volume~15 of {\em Proceedings of Machine Learning Research}, pages
  164--172, Fort Lauderdale, FL, USA, 11--13 Apr 2011. JMLR Workshop and
  Conference Proceedings.

\bibitem{bietti_mairal_invariance_stability}
Alberto Bietti and Julien Mairal.
\newblock Invariance and stability of deep convolutional representations.
\newblock In I. Guyon, U.~V. Luxburg, S. Bengio, H. Wallach, R. Fergus, S.
  Vishwanathan, and R. Garnett, editors, {\em Advances in Neural Information
  Processing Systems}, volume~30, pages 6210--6220. Curran Associates, Inc.,
  2017.

\bibitem{Bietti_mairal_group_invariance}
Alberto Bietti and Julien Mairal.
\newblock Group invariance, stability to deformations, and complexity of deep
  convolutional representations.
\newblock {\em J. Mach. Learn. Res.}, 20(1):876–924, Jan. 2019.

\bibitem{Bruna_mallat_invariant_scattering}
J. {Bruna} and S. {Mallat}.
\newblock Invariant scattering convolution networks.
\newblock {\em IEEE Transactions on Pattern Analysis and Machine Intelligence},
  35(8):1872--1886, 2013.

\bibitem{7913730}
L. {Chen}, G. {Papandreou}, I. {Kokkinos}, K. {Murphy}, and A.~L. {Yuille}.
\newblock Deeplab: Semantic image segmentation with deep convolutional nets,
  atrous convolution, and fully connected crfs.
\newblock {\em IEEE Transactions on Pattern Analysis and Machine Intelligence},
  40(4):834--848, 2018.

\bibitem{chen2014semantic}
Liang-Chieh Chen, George Papandreou, Iasonas Kokkinos, Kevin Murphy, and Alan~L
  Yuille.
\newblock Semantic image segmentation with deep convolutional nets and fully
  connected crfs.
\newblock {\em arXiv preprint arXiv:1412.7062}, 2014.

\bibitem{Cheng_rotation_invariant_cnn}
G. {Cheng}, P. {Zhou}, and J. {Han}.
\newblock Learning rotation-invariant convolutional neural networks for object
  detection in vhr optical remote sensing images.
\newblock {\em IEEE Transactions on Geoscience and Remote Sensing},
  54(12):7405--7415, 2016.

\bibitem{group_equivariant_convolution_networks}
Taco Cohen and Max Welling.
\newblock Group equivariant convolutional networks.
\newblock In {\em International conference on machine learning}, pages
  2990--2999, 2016.

\bibitem{cubuk_auto_augment}
Ekin~D. Cubuk, Barret Zoph, Dandelion Mane, Vijay Vasudevan, and Quoc~V. Le.
\newblock Autoaugment: Learning augmentation strategies from data.
\newblock In {\em Proceedings of the IEEE/CVF Conference on Computer Vision and
  Pattern Recognition (CVPR)}, June 2019.

\bibitem{deformable_cnns}
Jifeng Dai, Haozhi Qi, Yuwen Xiong, Yi Li, Guodong Zhang, Han Hu, and Yichen
  Wei.
\newblock Deformable convolutional networks.
\newblock In {\em Proceedings of the IEEE International Conference on Computer
  Vision (ICCV)}, Oct 2017.

\bibitem{deng2009imagenet}
Jia Deng, Wei Dong, Richard Socher, Li-Jia Li, Kai Li, and Li Fei-Fei.
\newblock Imagenet: A large-scale hierarchical image database.
\newblock In {\em 2009 IEEE conference on computer vision and pattern
  recognition}, pages 248--255. Ieee, 2009.

\bibitem{cutout_regularization}
Terrance DeVries and Graham~W Taylor.
\newblock Improved regularization of convolutional neural networks with cutout.
\newblock {\em arXiv preprint arXiv:1708.04552}, 2017.

\bibitem{cyclic_symmetry_rotation_cnn}
Sander Dieleman, Jeffrey De~Fauw, and Koray Kavukcuoglu.
\newblock Exploiting cyclic symmetry in convolutional neural networks.
\newblock In {\em Proceedings of the 33rd International Conference on
  International Conference on Machine Learning - Volume 48}, ICML'16, page
  1889–1898. JMLR.org, 2016.

\bibitem{8038465}
S. {Dodge} and L. {Karam}.
\newblock A study and comparison of human and deep learning recognition
  performance under visual distortions.
\newblock In {\em 2017 26th International Conference on Computer Communication
  and Networks (ICCCN)}, pages 1--7, 2017.

\bibitem{landscape_of_spatial_robustness}
Logan Engstrom, Brandon Tran, Dimitris Tsipras, Ludwig Schmidt, and Aleksander
  Madry.
\newblock Exploring the landscape of spatial robustness.
\newblock volume~97 of {\em Proceedings of Machine Learning Research}, pages
  1802--1811, Long Beach, California, USA, 09--15 Jun 2019. PMLR.

\bibitem{fawzi_are_classifiers_invariant}
Alhussein Fawzi and Pascal Frossard.
\newblock Manitest: Are classifiers really invariant?
\newblock 2015.

\bibitem{FUKUSHIMA2003161}
Kunihiko Fukushima.
\newblock Neocognitron for handwritten digit recognition.
\newblock {\em Neurocomputing}, 51:161 -- 180, 2003.

\bibitem{geirhos2017comparing}
Robert Geirhos, David~HJ Janssen, Heiko~H Sch{\"u}tt, Jonas Rauber, Matthias
  Bethge, and Felix~A Wichmann.
\newblock Comparing deep neural networks against humans: object recognition
  when the signal gets weaker.
\newblock {\em arXiv preprint arXiv:1706.06969}, 2017.

\bibitem{geirhos_generalization_in_humans}
Robert Geirhos, Carlos R.~M. Temme, Jonas Rauber, Heiko~H. Sch\"{u}tt, Matthias
  Bethge, and Felix~A. Wichmann.
\newblock Generalisation in humans and deep neural networks.
\newblock In S. Bengio, H. Wallach, H. Larochelle, K. Grauman, N. Cesa-Bianchi,
  and R. Garnett, editors, {\em Advances in Neural Information Processing
  Systems}, volume~31, pages 7538--7550. Curran Associates, Inc., 2018.

\bibitem{goodfellow_measuring_invariances}
Ian Goodfellow, Honglak Lee, Quoc~V. Le, Andrew Saxe, and Andrew~Y. Ng.
\newblock Measuring invariances in deep networks.
\newblock In Y. Bengio, D. Schuurmans, J.~D. Lafferty, C.~K.~I. Williams, and
  A. Culotta, editors, {\em Advances in Neural Information Processing Systems
  22}, pages 646--654. Curran Associates, Inc., 2009.

\bibitem{explaining_harnessing_adverserial}
Ian Goodfellow, Jonathon Shlens, and Christian Szegedy.
\newblock Explaining and harnessing adversarial examples.
\newblock In {\em International Conference on Learning Representations}, 2015.

\bibitem{he_2016_resnet}
Kaiming He, Xiangyu Zhang, Shaoqing Ren, and Jian Sun.
\newblock Deep residual learning for image recognition.
\newblock In {\em Proceedings of the IEEE Conference on Computer Vision and
  Pattern Recognition (CVPR)}, June 2016.

\bibitem{augmix}
Dan Hendrycks*, Norman Mu*, Ekin~Dogus Cubuk, Barret Zoph, Justin Gilmer, and
  Balaji Lakshminarayanan.
\newblock Augmix: A simple method to improve robustness and uncertainty under
  data shift.
\newblock In {\em International Conference on Learning Representations}, 2020.

\bibitem{warped_convolutions}
Jo\~{a}o~F. Henriques and Andrea Vedaldi.
\newblock Warped convolutions: Efficient invariance to spatial transformations.
\newblock In {\em Proceedings of the 34th International Conference on Machine
  Learning - Volume 70}, ICML'17, page 1461–1469. JMLR.org, 2017.

\bibitem{Hosseini_google_cloud}
H. {Hosseini}, B. {Xiao}, and R. {Poovendran}.
\newblock Google's cloud vision api is not robust to noise.
\newblock In {\em 2017 16th IEEE International Conference on Machine Learning
  and Applications (ICMLA)}, pages 101--105, 2017.

\bibitem{howard2017mobilenets}
Andrew~G Howard, Menglong Zhu, Bo Chen, Dmitry Kalenichenko, Weijun Wang,
  Tobias Weyand, Marco Andreetto, and Hartwig Adam.
\newblock Mobilenets: Efficient convolutional neural networks for mobile vision
  applications.
\newblock {\em arXiv preprint arXiv:1704.04861}, 2017.

\bibitem{ilyas2019adversarial}
Andrew Ilyas, Shibani Santurkar, Dimitris Tsipras, Logan Engstrom, Brandon
  Tran, and Aleksander Madry.
\newblock Adversarial examples are not bugs, they are features.
\newblock In {\em Advances in Neural Information Processing Systems}, pages
  125--136, 2019.

\bibitem{Amirul_jia_position_info}
Md~Amirul Islam*, Sen Jia*, and Neil D.~B. Bruce.
\newblock How much position information do convolutional neural networks
  encode?
\newblock In {\em International Conference on Learning Representations}, 2020.

\bibitem{DBLP:journals/corr/KanazawaSJ14}
Angjoo Kanazawa, Abhishek Sharma, and David~W. Jacobs.
\newblock Locally scale-invariant convolutional neural networks.
\newblock {\em CoRR}, abs/1412.5104, 2014.

\bibitem{Kanbak_2018_CVPR}
Can Kanbak, Seyed-Mohsen Moosavi-Dezfooli, and Pascal Frossard.
\newblock Geometric robustness of deep networks: Analysis and improvement.
\newblock In {\em Proceedings of the IEEE Conference on Computer Vision and
  Pattern Recognition (CVPR)}, June 2018.

\bibitem{cnn_absolute_position}
Osman~Semih Kayhan and Jan C.~van Gemert.
\newblock On translation invariance in cnns: Convolutional layers can exploit
  absolute spatial location.
\newblock In {\em Proceedings of the IEEE/CVF Conference on Computer Vision and
  Pattern Recognition (CVPR)}, June 2020.

\bibitem{krizhevsky2009learning}
Alex Krizhevsky, Geoffrey Hinton, et~al.
\newblock Learning multiple layers of features from tiny images.
\newblock 2009.

\bibitem{lecun2015deep_nature}
Yann LeCun, Yoshua Bengio, and Geoffrey Hinton.
\newblock Deep learning.
\newblock {\em nature}, 521(7553):436--444, 2015.

\bibitem{doi:10.1162/neco.1989.1.4.541}
Y. LeCun, B. Boser, J.~S. Denker, D. Henderson, R.~E. Howard, W. Hubbard, and
  L.~D. Jackel.
\newblock Backpropagation applied to handwritten zip code recognition.
\newblock {\em Neural Computation}, 1(4):541--551, 1989.

\bibitem{lecun_1990}
Yann LeCun, Bernhard~E. Boser, John~S. Denker, Donnie Henderson, R.~E. Howard,
  Wayne~E. Hubbard, and Lawrence~D. Jackel.
\newblock Handwritten digit recognition with a back-propagation network.
\newblock In D.~S. Touretzky, editor, {\em Advances in Neural Information
  Processing Systems 2}, pages 396--404. Morgan-Kaufmann, 1990.

\bibitem{lecun_gradient_based_learning}
Y. {Lecun}, L. {Bottou}, Y. {Bengio}, and P. {Haffner}.
\newblock Gradient-based learning applied to document recognition.
\newblock {\em Proceedings of the IEEE}, 86(11):2278--2324, 1998.

\bibitem{madry2018towards}
Aleksander Madry, Aleksandar Makelov, Ludwig Schmidt, Dimitris Tsipras, and
  Adrian Vladu.
\newblock Towards deep learning models resistant to adversarial attacks.
\newblock In {\em International Conference on Learning Representations}, 2018.

\bibitem{conv_kernel_networks}
Julien Mairal, Piotr Koniusz, Zaid Harchaoui, and Cordelia Schmid.
\newblock Convolutional kernel networks.
\newblock In Z. Ghahramani, M. Welling, C. Cortes, N. Lawrence, and K.~Q.
  Weinberger, editors, {\em Advances in Neural Information Processing Systems},
  volume~27, pages 2627--2635. Curran Associates, Inc., 2014.

\bibitem{mallat_group_invariant_scattering}
St{\'e}phane Mallat.
\newblock Group invariant scattering.
\newblock {\em Communications on Pure and Applied Mathematics},
  65(10):1331--1398, 2012.

\bibitem{manfredi2020shift}
Marco Manfredi and Yu Wang.
\newblock Shift equivariance in object detection.
\newblock {\em arXiv preprint arXiv:2008.05787}, 2020.

\bibitem{oppenheim2001discrete}
Alan~V Oppenheim, John~R Buck, and Ronald~W Schafer.
\newblock {\em Discrete-time signal processing. Vol. 2}.
\newblock Upper Saddle River, NJ: Prentice Hall, 2001.

\bibitem{ruderman2018pooling}
Avraham Ruderman, Neil~C Rabinowitz, Ari~S Morcos, and Daniel Zoran.
\newblock Pooling is neither necessary nor sufficient for appropriate
  deformation stability in cnns.
\newblock {\em arXiv preprint arXiv:1804.04438}, 2018.

\bibitem{Sifre_Mallat_Rotation_2013_CVPR}
Laurent Sifre and Stephane Mallat.
\newblock Rotation, scaling and deformation invariant scattering for texture
  discrimination.
\newblock In {\em Proceedings of the IEEE Conference on Computer Vision and
  Pattern Recognition (CVPR)}, June 2013.

\bibitem{simoncelli_shiftable_multiscale}
Eero~P Simoncelli, William~T Freeman, Edward~H Adelson, and David~J Heeger.
\newblock Shiftable multiscale transforms.
\newblock {\em IEEE transactions on Information Theory}, 38(2):587--607, 1992.

\bibitem{simoyan_vgg_2015}
Karen Simonyan and Andrew Zisserman.
\newblock Very deep convolutional networks for large-scale image recognition.
\newblock In Yoshua Bengio and Yann LeCun, editors, {\em 3rd International
  Conference on Learning Representations, {ICLR} 2015, San Diego, CA, USA, May
  7-9, 2015, Conference Track Proceedings}, 2015.

\bibitem{translation_insensitive_cnns}
Ganesh Sundaramoorthi and Timothy~E Wang.
\newblock Translation insensitive cnns.
\newblock {\em arXiv preprint arXiv:1911.11238}, 2019.

\bibitem{szegedy2013intriguing}
Christian Szegedy, Wojciech Zaremba, Ilya Sutskever, Joan Bruna, Dumitru Erhan,
  Ian Goodfellow, and Rob Fergus.
\newblock Intriguing properties of neural networks.
\newblock {\em arXiv preprint arXiv:1312.6199}, 2013.

\bibitem{VANNOORD2017583}
Nanne {van Noord} and Eric Postma.
\newblock Learning scale-variant and scale-invariant features for deep image
  classification.
\newblock {\em Pattern Recognition}, 61:583 -- 592, 2017.

\bibitem{vasiljevic2016blur_cnn}
Igor Vasiljevic, Ayan Chakrabarti, and Gregory Shakhnarovich.
\newblock Examining the impact of blur on recognition by convolutional
  networks.
\newblock {\em arXiv preprint arXiv:1611.05760}, 2016.

\bibitem{steerable_rotation_equivariant_cnn}
Maurice Weiler, Fred~A. Hamprecht, and Martin Storath.
\newblock Learning steerable filters for rotation equivariant cnns.
\newblock In {\em Proceedings of the IEEE Conference on Computer Vision and
  Pattern Recognition (CVPR)}, June 2018.

\bibitem{Worrall_2017_CVPR}
Daniel~E. Worrall, Stephan~J. Garbin, Daniyar Turmukhambetov, and Gabriel~J.
  Brostow.
\newblock Harmonic networks: Deep translation and rotation equivariance.
\newblock In {\em Proceedings of the IEEE Conference on Computer Vision and
  Pattern Recognition (CVPR)}, July 2017.

\bibitem{wu2015deep_image}
Ren Wu, Shengen Yan, Yi Shan, Qingqing Dang, and Gang Sun.
\newblock Deep image: Scaling up image recognition.
\newblock {\em arXiv preprint arXiv:1501.02876}, 7(8), 2015.

\bibitem{scale_invariant_cnn}
Yichong Xu, Tianjun Xiao, Jiaxing Zhang, Kuiyuan Yang, and Zheng Zhang.
\newblock Scale-invariant convolutional neural networks.
\newblock {\em arXiv preprint arXiv:1411.6369}, 2014.

\bibitem{Yu_2017_CVPR}
Fisher Yu, Vladlen Koltun, and Thomas Funkhouser.
\newblock Dilated residual networks.
\newblock In {\em Proceedings of the IEEE Conference on Computer Vision and
  Pattern Recognition (CVPR)}, July 2017.

\bibitem{Zhang19}
Richard Zhang.
\newblock Making convolutional networks shift-invariant again.
\newblock volume~97 of {\em Proceedings of Machine Learning Research}, pages
  7324--7334, Long Beach, California, USA, 09--15 Jun 2019. PMLR.

\bibitem{zou2020delving}
Xueyan Zou, Fanyi Xiao, Zhiding Yu, and Yong~Jae Lee.
\newblock Delving deeper into anti-aliasing in convnets.
\newblock {\em arXiv preprint arXiv:2008.09604}, 2020.

\end{thebibliography}
}

\newpage

\clearpage
\appendix

\section{Non-linear activation functions and shift invariance}
\label{sec:non_linear_ac}

We saw in Section \ref{sec:aps_motivation} of the paper that anti-aliasing a signal before downsampling restores sum-shift-invariance. In particular, consider a 1-D signal $x_0(n)$ and its 1-pixel shift $x_1(n) = x_0(n-1)$. Anti-aliasing the two signals (with an ideal low pass filter) followed by downsampling with stride 2 results in $y_0^a(n)$ and $y_1^a(n)$ with DTFTs
\begin{align}
Y_0^{a}(\w) = \frac{X_0(\w/2)}{2},\text{ }
\label{eq:poly_1}
Y_1^{a}(\w) = \frac{X_0(\w/2)e^{-j\w/2}}{2},
\end{align}
that satisfy $Y_0^{a}(0)  = Y_1^{a}(0)$. Azulay and Weiss pointed out in \cite{Azulay_Weiss} that the sum-shift invariance obtained via anti-aliasing is lost due to the action of non-linear activation functions like ReLU in convolutional neural networks. They postulated that this happens through the generation of high-frequency content after applying ReLU. We elaborate on this phenomenon here and also show that high frequencies alone do not provide a full picture. \\

\begin{figure}[t]
	\begin{center}
		\includegraphics[width=\linewidth]{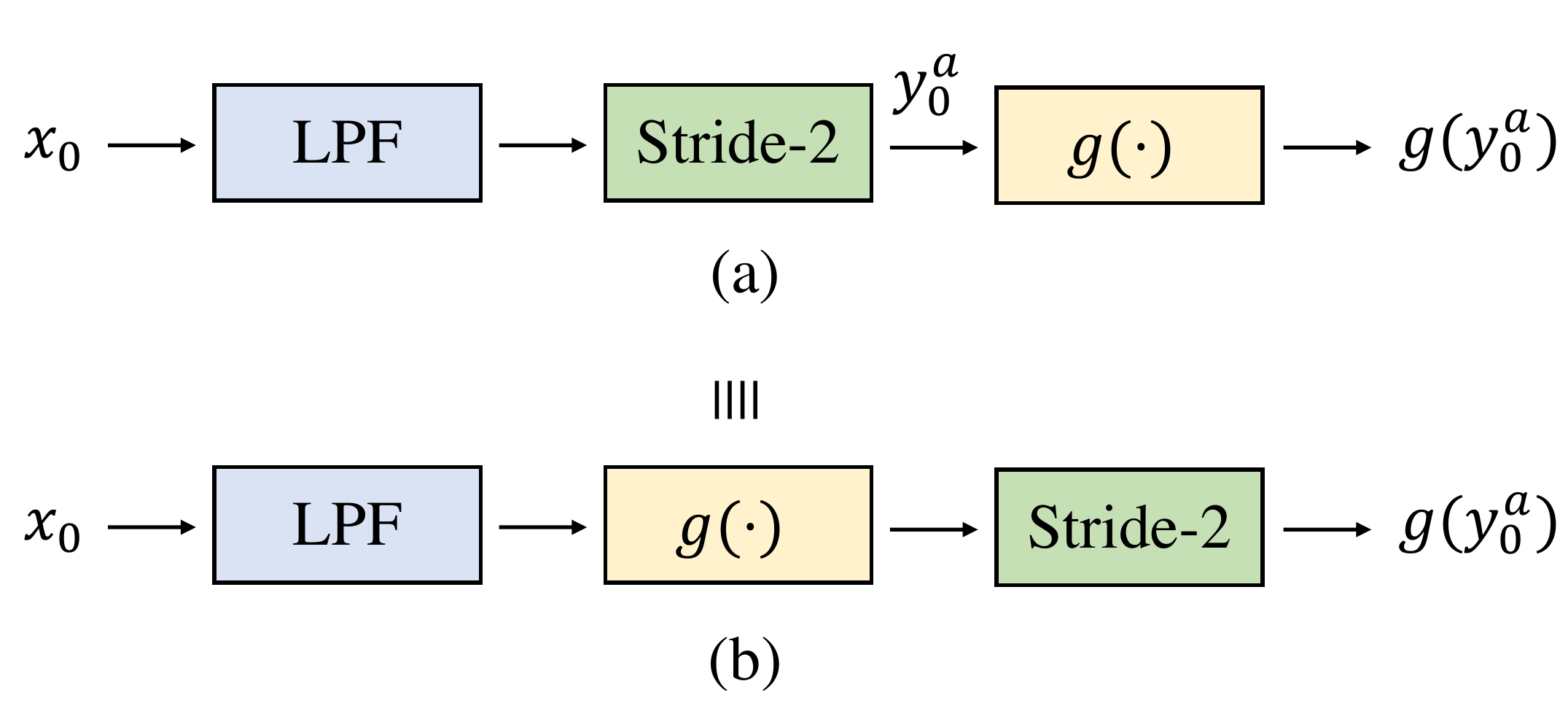}
	\end{center}
	\caption{Pointwise non-linearity $g$ can be interchanged with the stride operation. Despite anti-aliasing $x_0$ with LPF block, $g$ generates high frequencies which can lead to additional aliasing during downsampling.} 
	\label{fig:non_linearity_permute_block_diag}
\end{figure}
\begin{figure}
	\begin{center}
		\includegraphics[width=\linewidth]{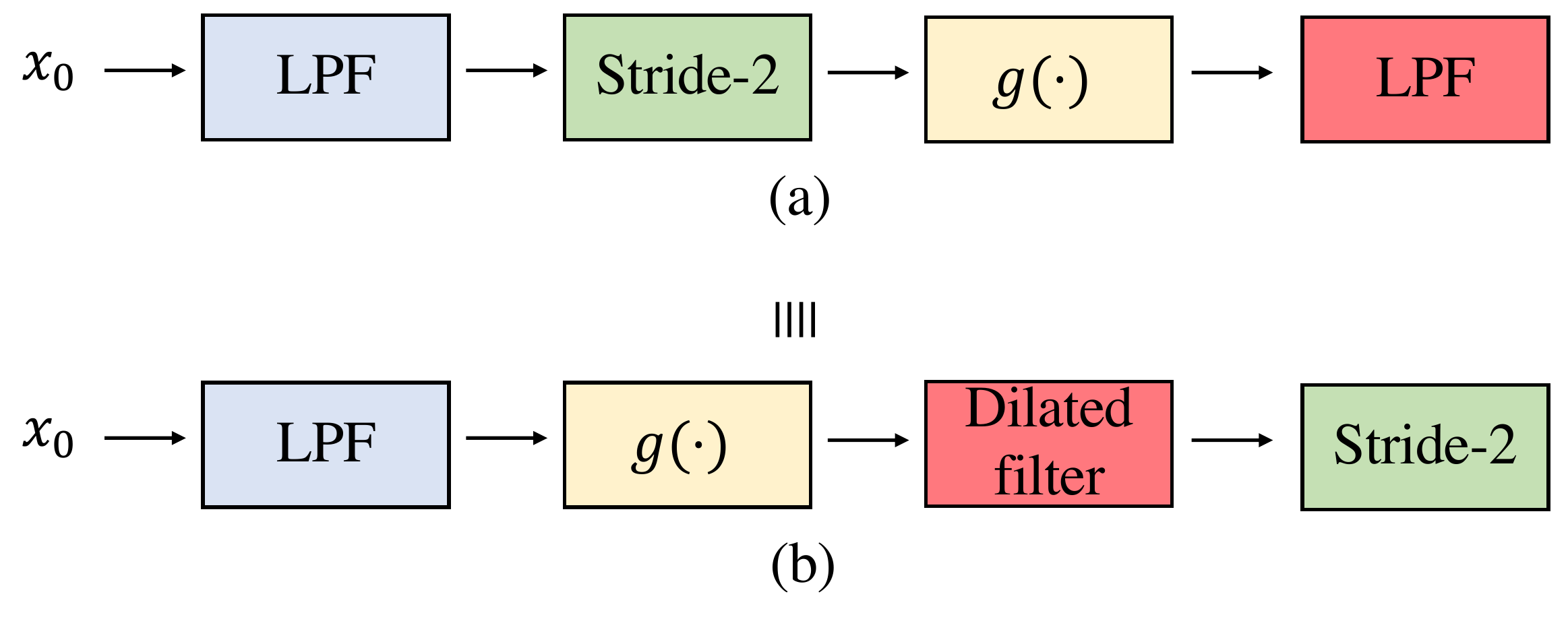}
	\end{center}
	\caption{Additional low pass filtering after $g$ in (a) does not eliminate the impact of aliasing. This is because, as shown in (b), interchanging the final LPF block with stride operation, results in a dilated version of the filter which is not low-pass any more.} 
	\label{fig:non_linearity_permute_block_diag_dil_lpf}
	
\end{figure}

Let $g(\cdot)$ be a generic pointwise non-linear activation function applied to the outputs of anti-aliased downsampling. Owing to the pointwise nature of $g$, the stride operation and the non-linearity can be interchanged, making the network block in Fig. \ref{fig:non_linearity_permute_block_diag}(a) equivalent to the one in Fig. \ref{fig:non_linearity_permute_block_diag}(b). Notice in Fig. \ref{fig:non_linearity_permute_block_diag}(b) that despite anti-aliasing $x_0$ with an ideal low pass filter LPF, $g$ generates additional high frequencies which can result in aliasing on downsampling. One can not simply use another low pass filter to get rid of these newly generated aliased components. For example, a new low pass filter block added after $g$ in Fig. \ref{fig:non_linearity_permute_block_diag_dil_lpf}(a) can be interchanged with the stride operation to result in a dilated filter which is not low pass any more (Fig. \ref{fig:non_linearity_permute_block_diag_dil_lpf}(b)). \\



While high frequencies generated by non-linear activations can lead to invariance loss for various choices of $g$, we show in Section \ref{sec:polynomial_non_lin} that this might not always be necessary. For example, polynomial activations, despite generating aliased components, do not impact sum-shift-invariance. Therefore, in addition to its high frequency generation ability, we also take a closer look at how the ReLU non-linearity affects sum-shift invariance in terms of its thresholding behavior in Section \ref{sec:relu_explanation}.

\subsection{Action of polynomial non-linearities on sum-shift invariance}
\label{sec:polynomial_non_lin}

In Theorem 1 from Section \ref{sec:aps_motivation} in the paper, we stated that for any integer $m>1$, non-linear activation functions of the form $g(y) = y^m$ do not impact sum-shift-invariance. We provide the proof below.




%


%

%

\begin{proof}
	Let the DTFTs of $z_0 = g(y_0^a)$ and $z_1 = g(y_1^a)$ be $Z_0(\w)$ and $Z_1(\w)$. Then by definition of the DTFT,
	\begin{align}
	\label{eq:sum_z0_z1}
	Z_0(0) = \sum_{n\in \Z}z_0(n) , \text{ and }Z_1(0) = \sum_{n\in \Z}z_1(n).
	\end{align}
	
	Since $z_0 = (y_0^a)^m$, and $z_1 = (y_1^a)^m$, we have
	
	\begin{align}
	Z_0(\w) = \Big( \underbrace{Y_0^a(\w) \circledast Y_0^a(\w) \circledast \ldots\circledast Y_0^a(\w)}_{\text{m times}}      \Big),\\
		Z_1(\w) = \Big( \underbrace{Y_1^a(\w) \circledast Y_1^a(\w) \circledast \ldots\circledast Y_1^a(\w)}_{\text{m times}}      \Big),
	\end{align}

		where $\circledast$ represents circular convolution. For $i \in \{0,1\}$, we can write 
		
	\begin{align}
	&Z_i(\w) = \notag\\ 
	\label{eq:z_i_defn}
	&\Big(\frac{1}{2\pi}\Big)^{m-1} \mulint Y_i^a(\alpha_1) \cdots  Y_i^a(\w - \sum_{i=1}^{m-1} \alpha_i)d\bar{\alpha},
	\end{align}
	
		where $\bar{\alpha} = (\alpha_1, \alpha_2,\ldots \alpha_{m-1})$. From \eqref{eq:poly_1}, we have
		\begin{equation}
		\label{eq:y1_y0_relation}
		Y_1^{a}(\w) = Y_0^{a}(\w)e^{-j\omega/2}. 
		\end{equation}
		
		Using \eqref{eq:z_i_defn} and \eqref{eq:y1_y0_relation}, we can write  $Z_1(\w) = Z_0(\w) e^{-\frac{j\w}{2}} $, which when combined with \eqref{eq:sum_z0_z1} gives  
				\begin{equation}
		\sum_{n\in \Z}z_0(n) = \sum_{n\in \Z}z_1(n).
		\end{equation}
		\end{proof}
		Using linearity of Fourier transform, the result in Theorem 1 can be extended to arbitrary polynomial activation functions of the form $g(y) = \sum _{i=0}^m a_i y^i$ with $m>1$.\\

%
%
%
%
%


\subsection{ReLU spoils sum-shift-invariance}
\label{sec:relu_explanation}
We now consider the ReLU non-linear activation function, $h(y) = \mathrm{relu}(y)$, which clips all negative values of signal $y$ to zero. Unlike the case with polynomials in Section \ref{sec:polynomial_non_lin}, deriving a closed form expression for the DTFTs of $h(y_0^a)$ and $h(y_1^a)$, for arbitrary $x_0$ and $x_1$ is non-trivial. We therefore analyze a simpler case where $x_0$ is assumed to be a cosine signal, and illustrate how sum-shift invariance is lost due to ReLUs.\\

Let $x_0$ be an $N$ length 1-D cosine and $x_1 = x_0(n-1)$ be its 1-pixel shift. We define the two signals as
\begin{align}
x_0 = \cos\Big(\frac{2\pi n }{N}\Big), \text{ and }x_1 = \cos\Big(\frac{2\pi (n-1) }{N}\Big)\\
n\in \{0, 1, \ldots N-1\}.\notag
\end{align}

 For any $N>4$, $x_0$ satisfies the Nyquist criterion and is anti-aliased by default. For $N' = N/2$ and $n\in \{0, 1, \ldots N'-1\}$, the downsampled outputs $y_0^a$ and $y_1^a$ are then given by
\begin{align}
y_0^a(n) = x_0(2n) = \cos&\Big(\frac{2\pi n }{N'}\Big),\\
y_1^a(n) = x_1(2n) = \cos&\Big(\frac{2\pi (n-1/2) }{N'}\Big).
\end{align}
Note that $y_0^a$ and $y_1^a$ are structurally similar signals, and can be interpreted as \textit{half-pixel} shifted versions of each other. The action of $h$ on $y_i^a$ can be regarded as multiplication by a window which is zero for any $ n$ where  $y_i^a(n)<0$. We construct sets $\{S_i^+\}_{i=0}^1$ containing $n$ where $y_i^a(n)>0$. For simplicity in constructing the sets, we assume $N'>6$ and divisible by 4 (similar conclusions from below can be reached without these simplifying assumptions as well). Then we have
\begin{align}
S_0^+ = &\Big\{n: n\in \Z, n\in  \Big[0, \frac{N'}{4}-1\Big] \cup \Big[\frac{3N'}{4}+1 , {N'}-1\Big] \Big\},\\
S_1^+ = &\Big\{n: n\in \Z, n\in \Big [0, \frac{N'}{4}\Big] \cup\Big[\frac{3N'}{4}+1 , {N'}-1 \Big] \Big\}.
\end{align}
Notice that the supports $S_0^+$ and $S_1^+$ differ by 1 pixel near $n=\frac{N'}{4}$. This is because despite being structurally similar, $y_0^a$ and $y_1^a$ have slightly different zero crossings, which results in some differences in the support of thresholded outputs. We can now compute the sums $\sum h(y_0^a)$ and $\sum h(y_1^a)$.
\begin{align}
\sum_{n \in \Z} h(y_0^a)(n) =& \sum_{n \in S_0^+}\cos\Big(\frac{2\pi n }{N'}\Big)\\
\label{eq:g_y0_def_1}
=&\text{ } \oRe\Big(  \sum_{n \in S_0^+}  e^{j\frac{2\pi n }{N'}}        \Big)\\
\label{eq:g_y0_def_2}
=& \text{ } \frac{\cos(2\pi/N)}{\sin(2\pi/N)}.
\end{align}


Similarly, $\sum_{n \in \Z} h(y_1^a)(n)$ is given by 

\begin{align}
\sum_{n \in \Z} h(y_1^a)(n) =&\text{ } \oRe \Big(  \sum_{n \in S_1^+}  e^{\frac{j2\pi (n-1/2)}{N'}}  \Big) \\
\label{eq:g_y1_def}
=&\text{ } \oRe \Big(e^{-\frac{j\pi}{N'}}\sum_{n \in S_1^+}  e^{j\frac{2\pi n }{N'}}        \Big).
\end{align}

We can rewrite \eqref{eq:g_y1_def} in terms of \eqref{eq:g_y0_def_1}, and get

\begin{align}
\sum_{n \in \Z} h(y_1^a)(n) & \\
=\text{ } \oRe \Big(e^{-\frac{j\pi}{N'}}\sum_{n \in S_0^+}  e^{j\frac{2\pi n }{N'}}  +& e^{-\frac{j\pi}{N'}}e^{\frac{j2\pi n }{N'}}|_{n=N'/4}       \Big)\\
\label{eq:sum_shift_invar_lost}
=\cos\Big(\frac{2\pi}{N}\Big)\sum_{n\in \Z} h(y_0^a)(n) +& \sin\Big(\frac{2\pi}{N}\Big).
\end{align}

%


\eqref{eq:sum_shift_invar_lost} illustrates the loss in sum-shift-invariance caused by ReLU. Notice that the differences in $\sum h(y_0^a)$ and $\sum h(y_1^a)$ arise due to minor differences in the signal content in $y_0^a$ and $y_1^a$, which are amplified by ReLU. The term $sin(2\pi/N)$ arises due to a 1-pixel difference in the supports of $h(y_0^a)$ and $h(y_1^a)$, whereas the cosine term is associated with $e^{-j\w/2}$ from \eqref{eq:poly_1}, again depicting the impact of small differences in $y_0^a$ and $y_1^a$.


\begin{figure}[t]
	\begin{center}
		\includegraphics[width=0.8\linewidth]{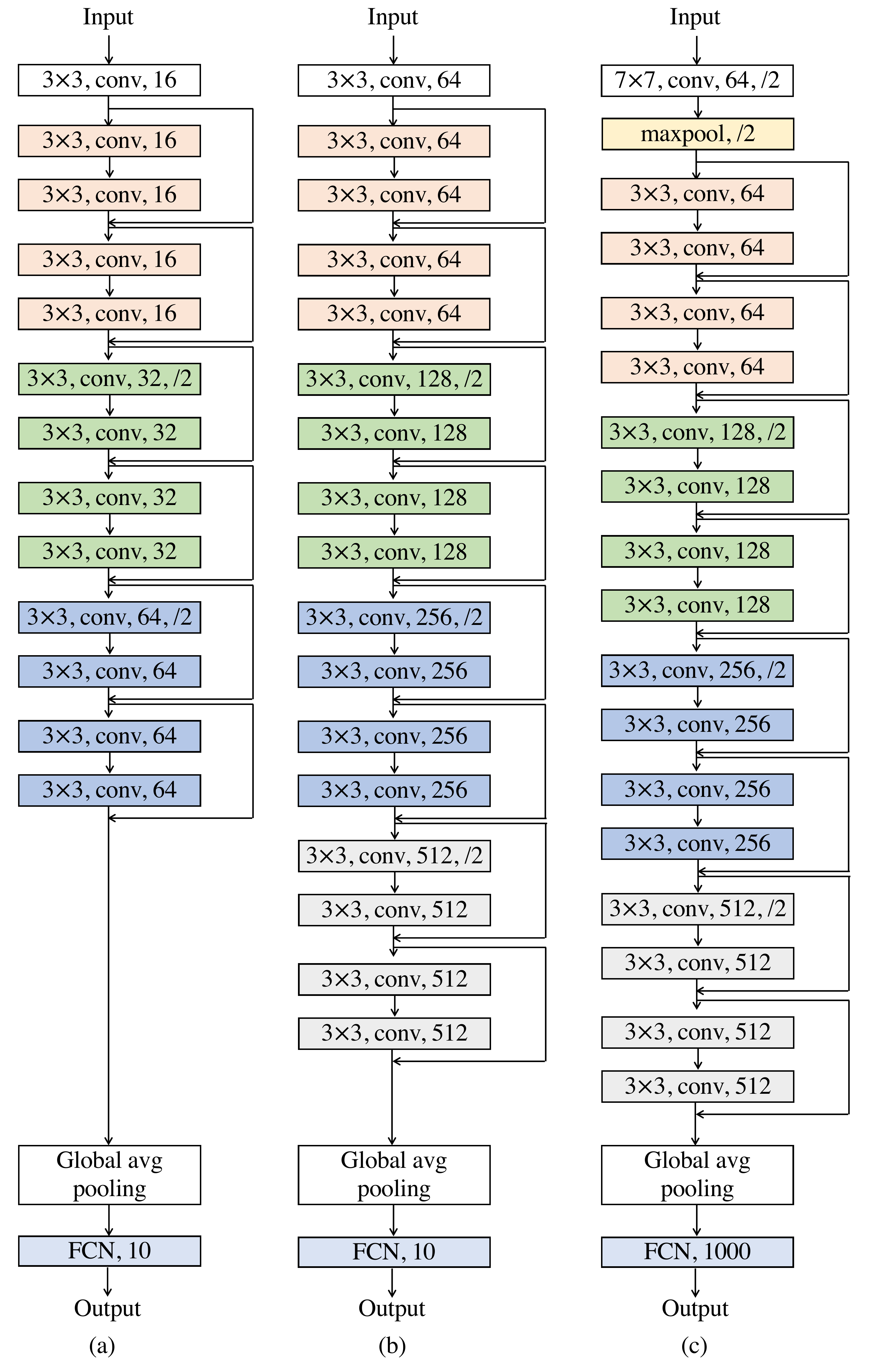}
	\end{center}
	\caption{Illustration of baseline ResNet architectures used in our experiments. (a) ResNet-20, (b) ResNet-18 used in CIFAR-10 classification. (c) Baseline ResNet-18 used in the ImageNet classification experiments.} 
	\label{fig:resnet_arch}
\end{figure}

\section{Implementation details}

We trained ResNet models with APS, anti-aliasing and baseline conventional downsampling approaches on CIFAR-10 and ImageNet datasets, and compared their achieved classification consistency and accuracy. For CIFAR-10 experiments, four variants of the architecture were used: ResNet-20, 56, 18 and 50. ResNet 20 and 56 were originally introduced in \cite{he_2016_resnet} for CIFAR-10 classification and are smaller models with number of channels: $\{16, 32, 64\}$ in different layers, and use stride 2 twice, which results in a resolution of $8\times 8$ in the final convolutional feature maps. On the other hand, ResNet-18 and 50 contain $\{64, 128, 256, 512\}$ number of channels, and downsample three times with a stride 2, resulting in final feature map resolution of $4\times 4$. Similar to the experiments with CIFAR-10 in \cite{he_2016_resnet}, we use a convolution with stride 1 and kernel size of $3\times3$ in the first convolutional layer. In all architectures, global average pooling layers are used at the end of the convolutional part of the networks. Fig. \ref{fig:resnet_arch}(a)-(b) illustrate the baseline architectures of ResNet-20 and 18 used in our experiments.\\

The original training set of the CIFAR-10 dataset was split into training and validation subsets of size 45k and 5k. All models were trained with batch size of 256 for 250 epochs using stochastic gradient descent (SGD) with momentum $0.9$ and weight decay $5\mathrm{e}{-4}$. The initial learning rate was chosen to be 0.1 and was decayed by a factor of 0.1 every 100 epochs. Training was performed on a single NVIDIA V-100 GPU. All the models were randomly initialized with a fixed seed before training. The models with the highest validation accuracy were used for evaluation on the test set.\\

For ImageNet classification, we used standard ResNet-18 model as baseline whose architecture is illustrated in Fig. \ref{fig:resnet_arch}(c). In all experiments, input image size of $224\times 224$ was used. The models were trained with batch size of 256 for 90 epochs using SGD with momentum 0.9 and weight decay of $1\mathrm{e}{-4}$. An initial learning rate of $0.1$ was chosen which was decayed by a factor of $0.1$ every 30 epochs. The models were trained in parallel on four NVIDIA V-100 GPUs. We report results for models with the highest validation accuracy.\\

 We were able to show significant improvements in consistency and accuracy with APS over baseline and anti-aliased downsampling without substantial hyper-parameter tuning. Further improvements in the results with better hyper-parameter search are therefore possible.

\subsection{Embedding APS in ResNet architecture}
\begin{figure}[t]
	\begin{center}
		\includegraphics[width=\linewidth]{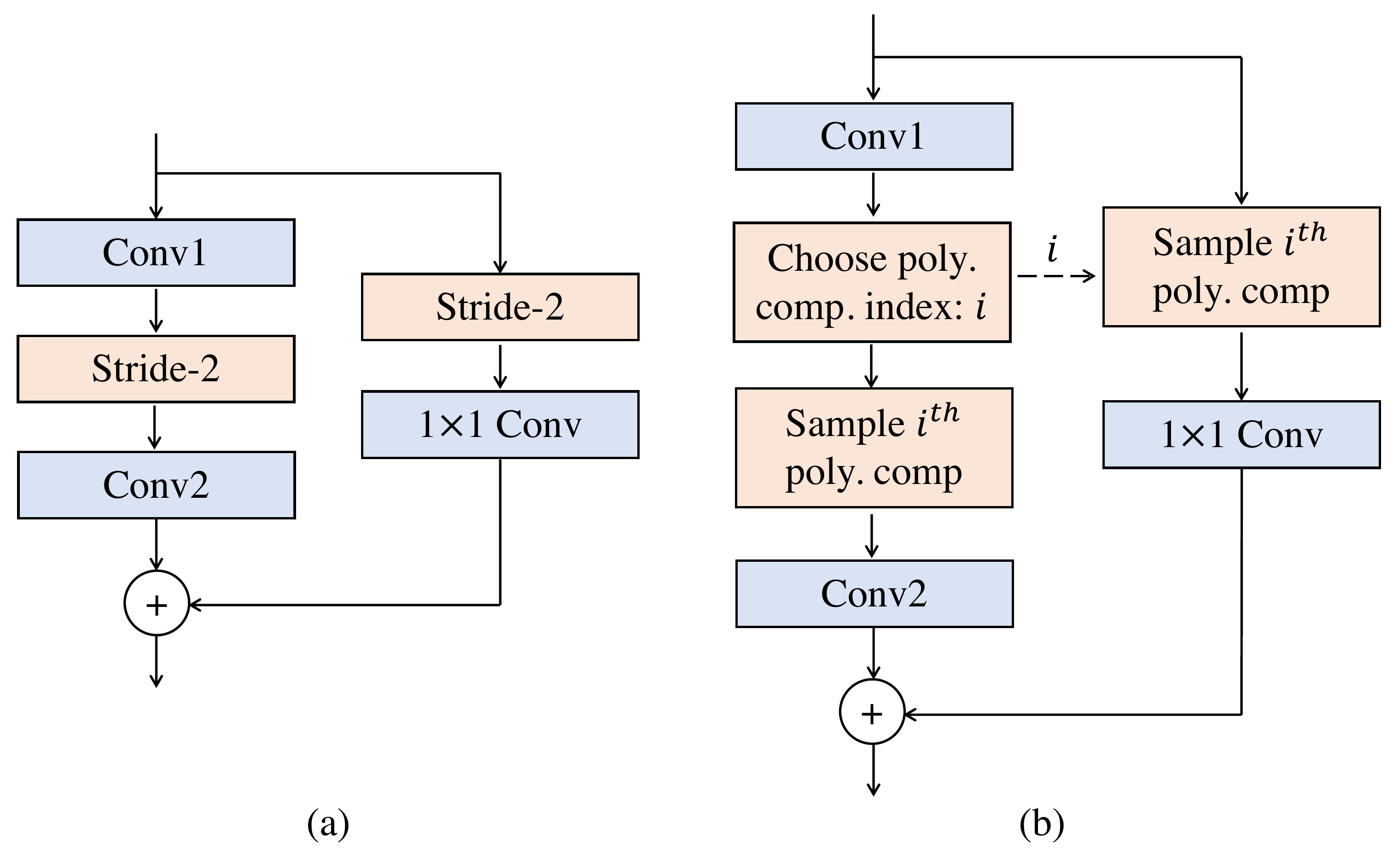}
	\end{center}
	\caption{Residual connection block with (a) baseline stride, (b) APS layer.} 
	\label{fig:integrating_aps_into_res_block}
\end{figure}

We replace the baseline stride layers in the ResNet architectures with APS modules. To ensure shift invariance, a consistent choice of polyphase components in the main and residual branch stride layer is needed. APS uses a permutation invariant criterion (like $\mathrm{argmax}$) to choose the component to be sampled in the main branch. The index of the chosen component is passed to the residual branch where the polyphase component with the same index is sampled. An illustration is provided in Fig. \ref{fig:integrating_aps_into_res_block}.

\section{Impact of polyphase component selection method on classification accuracy}

In the paper, we saw that APS achieves perfect shift invariance by selecting the polyphase component with the highest $l_2$ norm, i.e.
\begin{gather}
\label{eq:aps_criterion2}
\yaps = y_{i_1j_1},\\
\text{where }\text{ } i_1, j_1 = \argmax_{i,j} \{ \|y_{ij} \|_2  \}_{i,j=0}^1.\notag
\end{gather}

This can also be achieved, however, with other choices of shift invariant criteria. Here, we study the impact of different such criteria on the accuracy obtained on CIFAR-10 classification. In particular, we explore maximization of $l_p$ norms with $p=1$ and $\infty$ in addition to $p=2$. We also consider minimization of $l_1$ and $l_2$ norms. We run the experiments on ResNet-18 architecture with $9$ different initial random seeds and report the mean and standard deviation of achieved accuracy on the test set. \\

Table \ref{tab:aps_criterion_results} shows that choosing the polyphase component with the largest $l_\infty$ norm provides the highest classification accuracy which is then followed by choosing the one with the highest $l_2$ norm and $l_1$ norm. Additionally, the accuracy obtained when choosing polyphase component with minimum $l_2$ norm is somewhat lower than the case which chooses maximum $l_2$ norm. We believe this could be due to the polyphase components with higher energy containing more discriminative features.\\

Note that for all cases in Table \ref{tab:aps_criterion_results}, the achieved classification accuracy is $\sim 2\%$ higher than that of baseline ResNet-18 (reported in the paper). This is because in each case, APS enables stronger generalization via perfect shift invariance prior.






\begin{table}[t]
	\centering
	\begin{tabular}{c|c|c}
		\toprule
		APS criterion  &  Accuracy & Consistency\\
		\hline
		
		$\mathrm{argmax}$ $(l_1)$  &  $93.89\pm$ $0.27\%$  & \textbf{100\%}\\
$\mathrm{argmax}$ $(l_2)$  &  $94.03\pm$ $0.26\%$  & \textbf{100\%}\\
	$\mathrm{argmax}$ $(l_\infty)$  &  $\boldsymbol{94.14\pm}$ $\boldsymbol{0.25\%}$  & \textbf{100\%}\\
	\hline
	$\mathrm{argmin}$ $(l_1)$  &  $93.92\pm$ $0.12\%$  & \textbf{100\%}\\
	$\mathrm{argmin}$ $(l_2)$  &  $93.90\pm$ $0.16\%$  & \textbf{100\%}\\
		
		\bottomrule

	\end{tabular}
	
	\vspace{2pt}
	\caption{Impact of polyphase component selection method used by APS on CIFAR-10 classification accuracy.}
	\label{tab:aps_criterion_results}
\end{table}

\begin{table*}[t]
	\centering
	\begin{tabular}{c|*2c|*2c}
		\toprule
		\multicolumn{3}{c|}{Accuracy (unshifted)} & \multicolumn{2}{c}{Consistency}\\
		\hline
		Model  &  ResNet-18 & ResNet-50  &  ResNet-18 & ResNet-50 \\
		\hline
		Baseline &  91.96\%  &  90.05\%  & 90.88\%  & 88.96\% \\
		APS-3 &      94.53\%  &  93.80\%  & \textbf{100\%}  & \textbf{100\%} \\
		
		\hline
		
		Baseline + DA &   94.33\%        &   \textbf{94.77\% }   &      97.84\%     &   97.64\%      \\
		APS-3  + DA&      \textbf{94.61\%}   &      94.39\%     &   \textbf{100\% }      & \textbf{100\%}  \\

		\bottomrule

	\end{tabular}
	
	\vspace{2pt}
	\caption{Impact of APS on classification consistency and accuracy (evaluated on unshifted images) obtained using models trained with random shifts in data augmentation. Models trained without data augmentation are also shown for reference.}
	\label{tab:data_aug}
\end{table*}

\section{Experiments with data augmentation}

We saw in Section \ref{sec:classification_consistency_and_acc_test} of the paper that APS results in $100\%$ classification consistency and more than $2\%$ improvement in accuracy on CIFAR-10 dataset for models trained without any random shifts (data augmentation). Here, we assess how baseline sampling compares with APS when the models are trained on CIFAR-10 dataset with data augmentation (labelled as DA).  The results are reported in Table \ref{tab:data_aug}. \\

We observe that while data augmentation does improve classification consistency for baseline models, it is still lower than APS which yields perfect shift invariance. Classification accuracy, on the other hand, for both the baseline and APS is comparable (within the limits of training noise) when the models are trained with random shifts. This is not surprising because data augmentation is known to improve classification accuracy on images with patterns similar to the ones seen in training set. Note that, as reported in Section \ref{sec:out_of_dist_exps} of the paper, accuracy of networks with APS is more robust to image corruptions, and the models continue to yield $100\%$ classification consistency on all image distributions.

%

\section{Downsampling circularly shifted images with odd dimensions}
\label{sec:circ_shifted_images_odd}

\begin{figure}[t]
	\begin{center}
		\includegraphics[width=\linewidth]{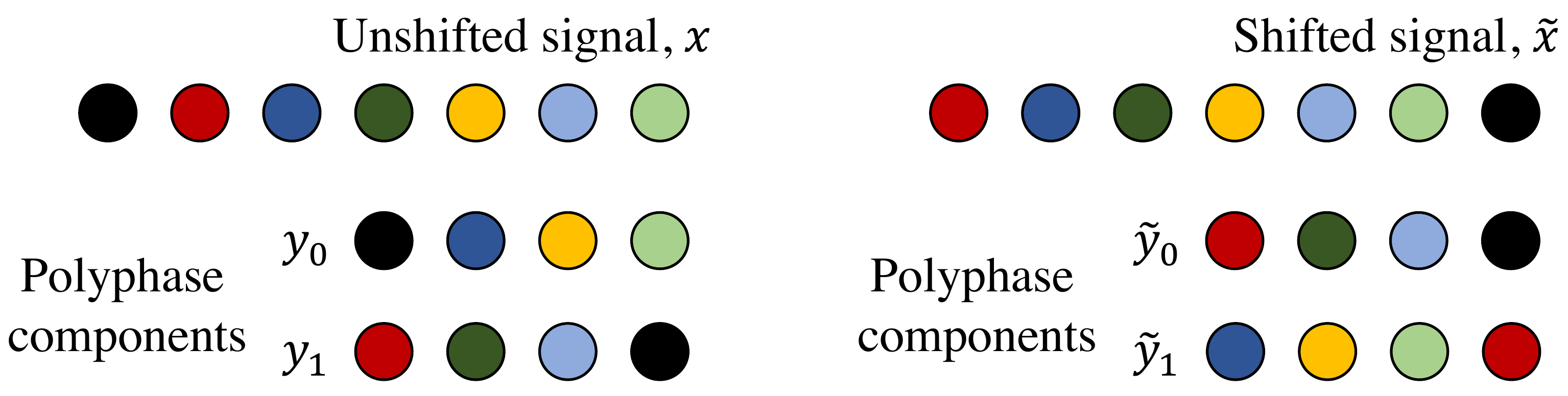}
	\end{center}
	\caption{Boundary artifacts associated with circular shifts. Downsampling an odd length signal and its circular shift can result in minor differences in polyphase components near their boundaries.} 
	\label{fig:odd_size_polyphase}
\end{figure}

With circular shift, pixels that exit from one end of a signal roll back in from the other, thereby preventing any information loss. While this makes circular shifts convenient for evaluating the impact of downsampling on shift invariance over finite length signals, they can lead to additional artifacts at the boundaries when sampling odd-sized signals. For example, as illustrated in Fig. \ref{fig:odd_size_polyphase}, while the polyphase components $y_1$ and $\tilde{y}_0$ are identical, $y_0$ and $\tilde{y}_1$ do not contain the same pixels near the boundaries. This is because downsampling an odd-sized signal with stride-2 breaks the periodicity associated with circular shifts, resulting in minor differences in the sets of polyphase components near the boundaries. \\

We investigate the impact of these artifacts by training ResNet-18 models with different downsampling modules on CIFAR-10 dataset with images center-cropped to size $30\times 30$. These images result in odd-sized feature maps inside the networks which generate boundary artifacts after downsampling. The models were then evaluated on $30\times 30$ center-cropped CIFAR-10 test set. Results in Table \ref{tab:cifar10_odd} show that despite the presence of artifacts, both the classification consistency and accuracy on unshifted images is greater for models that use APS.

\begin{table*}[t]
	\centering
	\begin{tabular}{c|c |c |c |c |c |c  |c | c  }
		\toprule
		& Baseline & APS & LPF-2&   APS-2 & LPF-3&   APS-3 & LPF-5&   APS-5 \\
		\hline
		Consistency&  88.23\%  &  98.13\%    &   94.28\%    &   98.48\%     &  96.15\%   &  98.65\%    &  98.02\%     &  \textbf{ 99.26\%} \\
		\hline
		Accuracy (unshifted images) &  90.91\%     &   93.99\%    &  93.22\%      &  93.83\%    &   93.56\%   &   \textbf{94.34\% }   &   94.28\%  &  94.22\% \\
		\bottomrule

	\end{tabular}
	
	\vspace{2pt}
	\caption{Classification consistency and accuracy obtained with different variants of ResNet-18 when evaluated on CIFAR-10 test set with images cropped to size $30\times 30$. The models were trained without seeing random shifts during training. Despite the presence of boundary effects caused by circular shifts on odd-sized feature maps, we observe higher consistency and accuracy with models containing APS. }
	\label{tab:cifar10_odd}
\end{table*}

\section{Timing analysis}
\label{sec:timing_analysis}

APS computes the norms of polyphase components for downsampling consistently to shifts. This leads to a modest increase in the time required to perform a forward pass in comparison with baseline network. For example, a forward pass on a $224\times224$ image with a circular padded baseline ResNet-18 takes $8.15\pm 0.47$ms on a single V-100 GPU. In comparison, ResNet-18 with APS layers takes $11.88\pm0.06$ms.

\end{document}